\documentclass[review]{elsarticle}
\usepackage{pgfplots}
\usepackage[utf8]{inputenc}
\usepackage{amsmath, nccmath}
\pgfplotsset{width=10cm,compat=1.8}
\usepackage{lineno,hyperref}
\usepackage[T1]{fontenc}
\usepackage{graphicx}
\usepackage{booktabs,multirow}
\usepackage{lipsum}
\usepackage{pgfplots,filecontents}
\usepackage{diagbox}
\usepackage{caption}
\usepackage[skip=0.5ex]{subcaption}
\usepackage{microtype}
\usepackage{tabularx}
\usepackage{graphicx}
\usepackage[export]{adjustbox}

\pgfplotsset{compat=newest}
\pgfplotsset{compat=1.16}
\usetikzlibrary{pgfplots.statistics}
\journal{Journal of \LaTeX\ Templates}

\makeatletter
\def\ps@pprintTitle{%
   \let\@oddhead\@empty
   \let\@evenhead\@empty
   \let\@oddfoot\@empty
   \let\@evenfoot\@oddfoot
}
\makeatother

\usepackage{times}
\usepackage{latexsym}
\usepackage{amsmath}
\usepackage{siunitx}
\usepackage{tikz}
\usepackage{array}
\usepackage{multirow}
\newcolumntype{G}{>{\centering\arraybackslash} m{2.6cm} }
\newcolumntype{H}{>{\centering\arraybackslash} m{1.7cm} }
\newcolumntype{C}{>{\centering\arraybackslash} m{0.65cm} }
\newcolumntype{P}{>{\centering\arraybackslash} m{2.05cm} } 
\newcolumntype{F}{>{\centering\arraybackslash} m{2.7cm} } 
\newcolumntype{E}{>{\centering\arraybackslash} m{1.1cm} } 
\newcolumntype{I}{>{\centering\arraybackslash} m{0.9cm} } 
\newcolumntype{L}{>{\centering\arraybackslash} m{4cm} }
\newcolumntype{M}{>{\centering\arraybackslash} m{2.7cm} }
\newcolumntype{N}{>{\centering\arraybackslash} m{2.6cm} }
\newcolumntype{B}{>{\left\arraybackslash} m{4cm} }

\begin{document}

\begin{frontmatter}
\title{Ensemble Learning-Based Approach for Improving Generalization Capability of Machine Reading Comprehension Systems}

\author[QOMaddress]{Razieh Baradaran}

\author[QOMaddress]{Hossein Amirkhani\corref{corauth1}}
\ead{amirkhani@qom.ac.ir}

\address[QOMaddress]{Department of Computer Engineering and Information Technology, University of Qom, Iran}

\cortext[corauth1]{Corresponding author:}





\begin{abstract}
Machine Reading Comprehension (MRC) is an active field in natural language processing with many successful developed models in recent years. Despite their high in-distribution accuracy, these models suffer from two issues: high training cost and low out-of-distribution accuracy. Even though some approaches have been presented to tackle the generalization problem, they have high, intolerable training costs. In this paper, we investigate the effect of ensemble learning approach to improve generalization of MRC systems without retraining a big model. After separately training the base models with different structures on different datasets, they are ensembled using weighting and stacking approaches in probabilistic and non-probabilistic settings. Three configurations are investigated including heterogeneous, homogeneous, and hybrid on eight datasets and six state-of-the-art models. We identify the important factors in the effectiveness of ensemble methods.
Also, we compare the robustness of ensemble and fine-tuned models against data distribution shifts. The experimental results show the effectiveness and robustness of the ensemble approach in improving the out-of-distribution accuracy of MRC systems, especially when the base models are similar in accuracies.
\end{abstract}


\end{frontmatter}

\nolinenumbers
\section{Introduction}

Machine Reading Comprehension (MRC) is one of the challenging and active tasks in Natural Language Processing (NLP) field. An MRC model is trained to extract or generate the \textit{answer} from a provided \textit{context} for a given \textit{question}. In recent years, with the progress of deep learning techniques, different neural MRC models have been developed \cite{RN238,yang2019xlnet,devlin2018bert}
surpassing human performance in some datasets like SQuAD \citep{RN22}. 

However, research shows that these models do not obtain a deep understanding of natural languages, and they are only able to provide acceptable answers for in-distribution questions and poorly generalize to unseen datasets. For example, the powerful and popular BERT model \cite{devlin2018bert} does not provide acceptable accuracies in out-of-distribution datasets \citep{talmor-berant-2019-multiqa}. This problem is due to the complexity and high diversity of natural languages that no single dataset can cover. Linzen ~\cite{linzen-2020-accelerate} blames the conventional evaluation paradigm, which he refers to as Pretraining-Agnostic Identically Distribution (PAID). This paradigm consists of evaluating a fine-tuned pre-trained model on a test set drawn from the same distribution as the training set. 


On the other hand, the current state-of-the-art MRC models have a very high, intolerable training cost. The computations required for deep learning research has experienced an estimated increase of 300,000x from 2012 to 2018, with training cost doubling every few months.\footnote{\url{https://openai.com/blog/ai-and-compute}} The side effect of this computation volume is a large carbon footprint which is environmentally unfriendly \citep{strubell2019energy}. In addition, this kind of research is considerably expensive, making barriers for institutes with limited resources. These issues have resulted in a new research focus to make AI both greener and more inclusive \citep{RN212}.


In this paper, we investigate the effectiveness of ensemble techniques to improve the generalization of MRC systems for out-of-distribution datasets. 
Instead of training a new model for every new dataset, we propose to ensemble the output of different trained models. This is similar to the way humans use to find the answers to their questions, where they aggregate the answers from different experts instead of seeking an omniscient.

More precisely, this paper examines the following questions:
\begin{itemize}
\item Is it possible to increase the accuracy of base models on a new dataset without re-training, only using the other models' predictions (zero-shot setting)?
\item With few available labeled data, what strategy does more robust generalization to unseen datasets, fine-tuning the base model or training an ensemble model (few-shot setting)?
\item What are the most important factors affecting the improvement rate of ensemble methods? 
\item Which model structure or training dataset plays a more effective role in creating more diverse base models?
\end{itemize}

To address the above questions,
we extend our previous preliminary work \cite{baradaran2021zero} by implementing several ensemble methods that can be categorized into weighting and stacking approaches. In the weighting approach, the outputs are aggregated based on some assigned weights. We study several weighting methods, ranging from simple equal weighting ensemble methods including mean, multiply, maximum, and minimum, to unequal weighting methods, which assign different weights to different base models. In the stacking approach, a new model learns how to weight and aggregate the outputs of different base models. 

Three different configurations are investigated in the experiments including different base models with the same training dataset (heterogeneous), the same model with different training datasets (homogeneous), and the hybrid configuration. 
We analyze the results of various ensemble methods in different configurations to illustrate the appropriate ensemble method in each situation. 
We also perform the robustness test for comparing the generalization capability of ensemble methods and the fine-tuning approach to unseen datasets.

The experiments indicate that the ensemble methods can improve generalization capability, especially in the hybrid setting in which the base models' accuracies are close to each other, without any labeled data. Also, with a few labeled data from the target dataset, we can train an ensemble model with robust performance against data distribution shift (unlike the fine-tuning approach).

 
Our main contributions are as follows:
\begin{itemize}
\item We propose the ensemble learning approach as a greener and more applicable way for improving the generalization of MRC models.
\item We investigate different ensemble methods for aggregating the outputs of MRC models including several weighting and stacking methods in probabilistic and non-probabilistic settings. Also, three configurations are experimentally investigated including heterogeneous, homogeneous, and hybrid settings. We analyze the results of different ensemble methods in different configurations and identify important success factors.
\item A robustness comparison between ensemble methods and the fine-tuning approach against data distribution shift is done.
\end{itemize}

The rest of this paper is organized as follows. In Section 2, related work are reviewed. The proposed approach is introduced in Section 3. The experiments are presented and discussed in Section 4, and the paper is concluded in the last section.

\section{Related work}
\subsection{Machine reading comprehension}
The history of MRC systems goes back to the 1970s. QUALM \cite{lehnert1977process} was one of the first MRC systems developed to answer questions about a story set, using simple hand-coded scripts. Lehnert emphasized that as human understanding of a story is proved by answering the questions about it, machine should also answer the questions about a given text to prove its understanding.

Due to the difficulty of the MRC task, very limited work were done in this field before 2013, especially before 2000; but with the success of deep learning methods and the development of large datasets, this task has again attracted the attention of researchers as one of the most popular tasks in natural language processing in recent years.
Attentive Reader \cite{hermann2015teaching} was the first neural MRC model based on a simple LSTM architecture and an attention mechanism trained on their collected CNN and Daily Mail dataset~\cite{hermann2015teaching}. It was able to obtain better results than the non-neural counterparts with a large margin.

Development of SQuAD~\citep{RN22} as the first large-scale MRC dataset in 2016 provided the necessary prerequisite to develop deeper models. BiDAF~\citep{RN2} was one popular neural network model introduced in 2017 with a bidirectional attention mechanism to extract two-way relations between the question and context.
With the introduction of more sophisticated models like RNet~\citep{wang2017r} and QANet~\citep{RN50}, the progress towards surpassing human-level performance in SQuAD was accelerated. So, further efforts were made to create more complex datasets, like HotpotQA\citep{RN244}, DROP\citep{RN206}, and RACE\citep{lai2017race}, that required more linguistic inferences and a deeper understanding of the text. 
In recent years, many studies have focused on developing models that can perform multi-step reasoning \citep{ hu2018reinforced,gong2018ruminating,bauer2018commonsense,yan2019deep,ding2019cognitive,jin2020mmm,zhang2020read}, answer non-factoid questions with free-form long answers~\citep{bauer2018commonsense,tan2018s,nishida2019multi,zhang2020sg}, and answers questions from multiple long passages instead of only a single paragraph~\citep{ clark2018simple,tu2019multi,tu2020select,ren2020multi}.

With the success of the Transformer architecture \citep{vaswani2017attention} in sequence modeling tasks, new powerful transformer-based language models like BERT~\cite{devlin2018bert} and XLNet \cite{yang2019xlnet} have  been proposed. These models are pre-trained and can be fine-tuned on different NLP tasks by adding some extra layers. They have obtained the best results not only in MRC, but also in several other NLP tasks~\cite{DCNplus-Zhang,su-etal-2019-generalizing, Alshahrani-2020, xiao-etal-2020-modeling, Gong-2019}.

However, more investigations indicate that these models suffer from the lack of generalization capability, so that their performance drops significantly in out-of-distribution datasets. Talmor and Berant \cite{talmor-berant-2019-multiqa} run experiments to analyze how well the MRC models generalize to unseen datasets, showing the poor generalization of these models even the successful BERT model. Sen and Saffari \cite{sen-saffari-2020} analyze BERT-based models trained on different datasets from various aspects including generalization, the reading comprehension ability, and robustness against question variations. They indicate that the available models suffer from lack of generalization and real comprehension, and rely on some simple heuristics to answer the input questions.

\subsection{Ensemble learning for improving generalization}
Ensemble learning is a set of methods that combine multiple base learners to make a better decision \cite{sagi2018ensemble}.
In this section, we review the papers intending to improve the generalization capability by ensemble learning.
Kim et al. \cite{RN250} investigated the effect of ensemble approach in domain adaptation for slot tagging and intent classification tasks. They focused on improving the generalization using $k$ pre-trained models on different domains. Then, a model is trained on the $(k+1)$-th domain using the attention weights of all $k$ models. This work is different from ours in two aspects. First, we do not train an MRC model for the new domain, but we train an ensemble model to aggregate different models' outputs. Second, we use different available model structures trained on different datasets so they have a diverse answering ability. 

Guo et al. \cite{RN251} proposed a domain adaptation model for multi-source settings using the mixture of experts (MoE) approach. They tested their method on sentiment analysis and POS tagging tasks. A new distance measure was proposed for calculating the distance between the given sample and each existing domain in an unsupervised way. The proposed model has a domain-invariant layer shared between domains for distance measurement. Obviously, this model is a complex one with many parameters which is in contrary with our purpose.

López et al. \cite{RN228} proposed an evolutionary-based ensemble technique to aggregate multiple models' outputs according to the input text domain in the sentiment analysis task. The ensemble models were trained on new datasets; however, in real-world settings, there may be no or little labeled data from the new domain.
Clark et al. \cite{clark-etal-2019-dont} introduced an ensemble-based two-stage method to improve the out-of-domain performance by avoiding dataset biases. They proposed to first train a naive model to make predictions only based on the biases, and then train a robust model in an ensemble with the bias-only one. 


There has also been some research on generalization capability of MRC models in out-of-distribution data. One of the recent attempts in this area is MRQA 2019 shared task \citep{RN230}, where 18 different MRC datasets are unified to develop models with high generalization capability. CLER \cite{RN231} is one of the developed model approaching MRQA task which is a large-scale model created based on multi-task learning, MoE approach \cite{RN227}, and ensemble techniques. It uses BERT \cite{devlin2018bert} as the shared layer and train natural language inference and MRC models in a multi-task setting. Above the BERT layer, it uses a MoE layer including $k$ experts to capture domain specific representations. Finally, to enhance generalization in test time, CLER has ensembled outputs of three independent models trained with different seeds. 
Even though this and many other methods proposed for MRQA task improve the out-of-distribution accuracy, they mainly suffer from high training costs which is in contrast to our aim of developing green systems. In the next section, we propose the details of our ensemble learning-based approach. 


\section{Proposed approach}
The neural MRC models can be categorized into extractive and abstractive modes based on their outputs \cite{RN249}. In the extractive mode, the output is an exact span in the context; while in the abstractive mode, the answer can be generated as a free-form text. Most research papers have focused on extractive mode due to the available datasets and its simplicity. In this paper, we also focus on ensembling the extractive MRC models.

In extractive neural MRC models, the final layer's output is usually two probability distributions over different tokens in the context, one distribution for the starting index $P_{s}=[p_{s_{1}},p_{s_{2}},...,p_{s_{l_{p}}}]$ and another for the end index $P_{e}=[p_{e_{1}},p_{e_{2}},...,p_{e_{l_{p}}}]$, where $l_{p}$ is the total number of tokens in the input context. The most probable answer span is extracted based on these output distributions.

In this section, we present the details of our ensemble learning-based approach for aggregating the outputs of different extractive neural MRC models. Ensemble learning is a set of methods to aggregate multiple models' outputs which is often used to improve the overall performance and generalization. These methods can be categorized into heterogeneous and homogeneous approaches. In the heterogeneous approach, the base models are built using a range of different learning algorithms; while in the homogeneous approach, all base models use the same learning algorithm \cite{RN255}.

We develop several ensemble methods in heterogeneous, homogeneous, and hybrid configurations to answer questions from new domains. In addition, probabilistic and non-probabilistic inputs to the ensemble module are investigated. In the probabilistic approach, the probabilistic outputs of the base models are used as inputs to the ensemble module; while in the non-probabilistic approach, the one-hot representation of start (end) vectors are used as ensemble's inputs, where the most probable start (end) tokens are set to one and the other tokens are zero. 

From another perspective, the ensemble learning includes the weighting methods \cite{RN256,RN261} that weight the base models' outputs, selection methods \cite{RN257,RN258} that select a subset of models to ensemble, and stacking methods \cite{RN259,RN260} that train a model on the outputs of base models to generate the final prediction. The selection methods can be considered as a special case of weighting where excluded models are assigned weight zero. 


\subsection{Weighting}
The simplest way to assign weights to the base models is to treat all models equally, which is called \textit{equal weighting}. On the other hand, the \textit{unequal weighting} assigns different weights to different base models to emphasize on more accurate models. 

\paragraph{Equal weighting}
As the simplest ensembling method, we treat all base models equally and use simple element-wise operations including mean, multiply, maximum, and minimum to aggregate the models' outputs as follows: 
\begin{equation}
\label{eq:simple}
\begin{aligned}
\text{Mean:}\hspace{15px}
P_{s}&=(P^{\{1\}}_{s}+...+P^{\{m\} }_{s})/m=\\&[(p^{\{1\}}_{s_{1}}+...+p^{\{m\}}_{s_{1}})/m,...,(p^{\{1\}}_{s_{l_{p}}}+...+p^{\{m\}}_{s_{l_{p}}})/m],
\\
\text{Multiply:}\hspace{15px}
P_{s}&\propto(P^{\{1\}}_{s}\times...\times P^{\{m\}}_{s})=\\&[(p^{\{1\}}_{s_{1}}\times...\times p^{\{m\}}_{s_{1}}),...,(p^{\{1\}}_{s_{l_{p}}}\times...\times p^{\{m\}}_{s_{l_{p}}})],
\\
\text{Maximum:}\hspace{15px}
P_{s}&\propto\max(P^{\{1\}}_{s},...,P^{\{m\}}_{s})=\\&[\max(p^{\{1\}}_{s_{1}},...,p^{\{m\}}_{s_{1}}),...,\max(p^{\{1\}}_{s_{l_{p}}},...,p^{\{m\}}_{s_{l_{p}}})],
\\
\text{Minimum:}\hspace{15px}
P_{s}&\propto\min(P^{\{1\}}_{s},...,P^{\{m\}}_{s})=\\&[\min(p^{\{1\}}_{s_{1}},...,p^{\{m\}}_{s_{1}}),...,\min(p^{\{1\}}_{s_{l_{p}}},...,p^{\{m\}}_{s_{l_{p}}})],
\end{aligned}
\end{equation}
where $m$ is the number of base models, $P^{\{i\}}_{s}$ is the start probability distribution of the $i$-th model, and $l_{p}$ is the length of the context. The end probability distribution is obtained in a similar way.

In \textit{mean} aggregation, a position is declared as the final answer if the arithmetic mean of models' outputs for that token is greater than others. The \textit{multiply} method considers the position where all models generate relatively high probabilities for it. It is similar to the Hinton's Product of Expert (PoE) technique \cite{hinton2002training}. The \textit{maximum} function is an optimistic method, because if just one model is highly confident about a position, it is considered as the correct output. On the other hand, the \textit{minimum} function is pessimistic, because it considers the output of the least-confident model as the label for a particular position. 

\paragraph{Unequal weighting}
In this approach, different models are treated differently based on their assigned weights:
\begin{equation}\label{eq:weighting1}
\begin{aligned}
&P_{s}\propto(w_{1}\times P^{\{1\}}_{s}+...+w_{m}\times P^{\{m\}}_{s}),\\
&P_{e}\propto(w_{1}\times P^{\{1\}}_{e}+...+w_{m}\times P^{\{m\}}_{e}),
\end{aligned}
\end{equation}
where $w_{i}$ is the weight of $i$-th model.

The problem here is how to determine the appropriate weights. We use the approach presented in \citep{RN261}, where the accuracy of a base model on a small set in the new domain is raised to the power of $\alpha$ to obtain its weight:
\begin{equation}\label{eq:weighting}
    \begin{aligned}
    &P_{s}\propto(acc^{\alpha}_{1}\times P^{\{1\}}_{s}+...+acc^{\alpha}_{m}\times P^{\{m\}}_{s}),\\
    &P_{e}\propto(acc^{\alpha}_{1}\times P^{\{1\}}_{e}+...+acc^{\alpha}_{m}\times P^{\{m\}}_{e}).
    \end{aligned}
\end{equation}
Finding a suitable value for the hyper-parameter $\alpha$ is important, because a very high value causes the result to converge to the strongest model and ignore the rest, while a very low value results in an equal contribution of different base models ignoring the actual weights. In Large et al. work~\citep{RN261}, this hyper-parameter
 is fixed for all experiments, which we refer it as \textit{fixed}. But in our work, we could not find a fixed value suitable for all settings.

We observe that if the base models have very different accuracies in a new domain, a low value of $\alpha$ causes the inaccurate models to destroy the outputs of stronger models. On the other hand, if the models have similar accuracies, a high value of $\alpha$ results in the selection of just one model with slightly higher accuracy and loosing the ensemble benefits. Based on these observations, we propose an \textit{automatic tuning} method to determine the value of $\alpha$ as the root of the difference between the two largest accuracies:
\begin{equation}\label{eq:auto}  
    \begin{aligned}
    \alpha=\left \lfloor{\sqrt{(acc_{largest}-acc_{sec\_largest})}}\right \rfloor,
    \end{aligned}
\end{equation}
where $acc_{largest}$ and $acc_{sec\_largest}$ are the largest and second-largest accuracies of base models, respectively. 

\subsection{Stacking}
In this approach, a new layer is responsible for aggregating the base models' outputs, which is trained using a training set from the new dataset. This layer produces the final probability distributions, getting the start and end probability distributions of base models as input:
\begin{equation}\label{eq:stacking}  
\begin{aligned}
&P_{s}=f_s(P^{\{1\}}_{s},P^{\{2\}}_{s},...,P^{\{m\}}_{s},P^{\{1\}}_{e},P^{\{2\}}_{e},...,P^{\{m\}}_{e}),\\
&P_{e}=f_e(P^{\{1\}}_{s},P^{\{2\}}_{s},...,P^{\{m\}}_{s},P^{\{1\}}_{e},P^{\{2\}}_{e},...,P^{\{m\}}_{e}),
\end{aligned}
\end{equation}
where $f_s$ and $f_e$ are the learned models. 

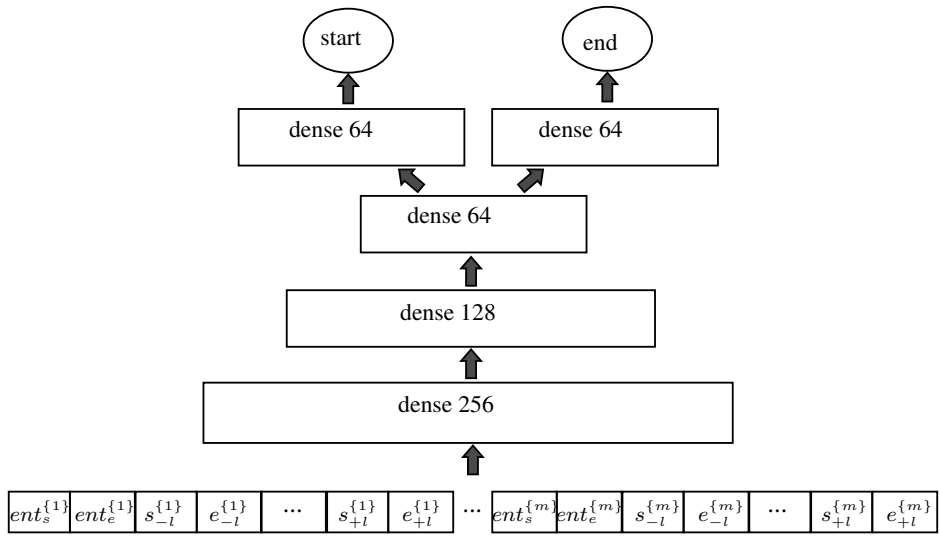
\begin{figure}[h]
\centering
\tikzset{every picture/.style={line width=0.70pt}} 

\begin{tikzpicture}[x=0.70pt,y=0.70pt,yscale=-1,xscale=1]

\draw   (182.3,207.4) -- (468.3,207.4) -- (468.3,240) -- (182.3,240) -- cycle ;
\draw   (227.3,157.4) -- (426.3,157.4) -- (426.3,188) -- (227.3,188) -- cycle ;
\draw   (267.3,106.4) -- (389.3,106.4) -- (389.3,137) -- (267.3,137) -- cycle ;
\draw   (338.3,59.4) -- (460.3,59.4) -- (460.3,90) -- (338.3,90) -- cycle ;
\draw   (201.3,59.4) -- (323.3,59.4) -- (323.3,90) -- (201.3,90) -- cycle ;
\draw  [fill={rgb, 255:red, 74; green, 74; blue, 74 }  ,fill opacity=1 ] (321.53,195.64) -- (326.91,189.4) -- (332.3,195.64) -- (329.61,195.64) -- (329.61,205) -- (324.22,205) -- (324.22,195.64) -- cycle ;
\draw   (376.39,21.46) .. controls (376.39,11.91) and (387.2,4.16) .. (400.54,4.16) .. controls (413.88,4.16) and (424.69,11.91) .. (424.69,21.46) .. controls (424.69,31.01) and (413.88,38.76) .. (400.54,38.76) .. controls (387.2,38.76) and (376.39,31.01) .. (376.39,21.46) -- cycle ;
\draw   (236.39,22.46) .. controls (236.39,12.91) and (247.2,5.16) .. (260.54,5.16) .. controls (273.88,5.16) and (284.69,12.91) .. (284.69,22.46) .. controls (284.69,32.01) and (273.88,39.76) .. (260.54,39.76) .. controls (247.2,39.76) and (236.39,32.01) .. (236.39,22.46) -- cycle ;
\draw  [fill={rgb, 255:red, 74; green, 74; blue, 74 }  ,fill opacity=1 ] (321.53,145.64) -- (326.91,139.4) -- (332.3,145.64) -- (329.61,145.64) -- (329.61,155) -- (324.22,155) -- (324.22,145.64) -- cycle ;
\draw  [fill={rgb, 255:red, 74; green, 74; blue, 74 }  ,fill opacity=1 ] (357.59,92.1) -- (365.83,92.12) -- (364.6,100.27) -- (362.85,98.23) -- (355.75,104.32) -- (352.24,100.23) -- (359.34,94.14) -- cycle ;
\draw  [fill={rgb, 255:red, 74; green, 74; blue, 74 }  ,fill opacity=1 ] (289.23,100.28) -- (287.98,92.13) -- (296.23,92.09) -- (294.48,94.14) -- (301.59,100.22) -- (298.09,104.31) -- (290.98,98.24) -- cycle ;
\draw  [fill={rgb, 255:red, 74; green, 74; blue, 74 }  ,fill opacity=1 ] (395.15,46) -- (400.54,39.76) -- (405.92,46) -- (403.23,46) -- (403.23,55.36) -- (397.84,55.36) -- (397.84,46) -- cycle ;
\draw  [fill={rgb, 255:red, 74; green, 74; blue, 74 }  ,fill opacity=1 ] (255.15,47) -- (260.54,40.76) -- (265.92,47) -- (263.23,47) -- (263.23,56.36) -- (257.84,56.36) -- (257.84,47) -- cycle ;
\draw   (145.62,266.34) -- (178.66,266.34) -- (178.66,288.36) -- (145.62,288.36) -- cycle ;
\draw   (178.66,266.34) -- (213.75,266.34) -- (213.75,288.36) -- (178.66,288.36) -- cycle ;
\draw   (213.35,266.34) -- (248.44,266.34) -- (248.44,288.36) -- (213.35,288.36) -- cycle ;
\draw   (249.13,266.34) -- (282.17,266.34) -- (282.17,288.36) -- (249.13,288.36) -- cycle ;
\draw   (282.17,266.34) -- (317.26,266.34) -- (317.26,288.36) -- (282.17,288.36) -- cycle ;
\draw   (77.08,266.34) -- (110.12,266.34) -- (110.12,288.36) -- (77.08,288.36) -- cycle ;
\draw   (110.12,266.34) -- (145.21,266.34) -- (145.21,288.36) -- (110.12,288.36) -- cycle ;
\draw   (408.71,266.34) -- (441.75,266.34) -- (441.75,288.36) -- (408.71,288.36) -- cycle ;
\draw   (441.75,266.34) -- (476.84,266.34) -- (476.84,288.36) -- (441.75,288.36) -- cycle ;
\draw   (477.25,266.34) -- (510.29,266.34) -- (510.29,288.36) -- (477.25,288.36) -- cycle ;
\draw   (510.67,266.34) -- (543.71,266.34) -- (543.71,288.36) -- (510.67,288.36) -- cycle ;
\draw   (543.71,266.34) -- (578.8,266.34) -- (578.8,288.36) -- (543.71,288.36) -- cycle ;
\draw   (338.35,266.34) -- (373.21,266.34) -- (373.21,288.4) -- (338.35,288.4) -- cycle ;
\draw   (373.21,266.34) -- (408.3,266.34) -- (408.3,288.36) -- (373.21,288.36) -- cycle ;
\draw  [fill={rgb, 255:red, 74; green, 74; blue, 74 }  ,fill opacity=1 ] (322.53,247.64) -- (327.91,241.4) -- (333.3,247.64) -- (330.61,247.64) -- (330.61,257) -- (325.22,257) -- (325.22,247.64) -- cycle ;

\draw (286,213) node [anchor=north west][inner sep=0.75pt]  [font=\small] [align=left] {dense 256};
\draw (287,163) node [anchor=north west][inner sep=0.75pt]  [font=\small] [align=left] {dense 128};
\draw (291,111) node [anchor=north west][inner sep=0.75pt]  [font=\small] [align=left] {dense 64};
\draw (362,64) node [anchor=north west][inner sep=0.75pt]  [font=\small] [align=left] {dense 64};
\draw (227,64) node [anchor=north west][inner sep=0.75pt]  [font=\small] [align=left] {dense 64};
\draw (386,17) node [anchor=north west][inner sep=0.75pt]  [font=\small] [align=left] {end};
\draw (244,16) node [anchor=north west][inner sep=0.75pt]  [font=\small] [align=left] {start};
\draw (148.62,269.34) node [anchor=north west][inner sep=0.75pt]  [font=\fontsize{0.78em}{0.94em}\selectfont]   {$s^{\{1\}}_{-l}$};
\draw (183.75,269.34) node [anchor=north west][inner sep=0.75pt]  [font=\fontsize{0.78em}{0.94em}\selectfont]   {$e^{\{1\}}_{-l}$};
\draw (253.48,269.34) node [anchor=north west][inner sep=0.75pt]  [font=\fontsize{0.78em}{0.94em}\selectfont]   {$s^{\{1\}}_{+l}$};
\draw (288.04,269.34) node [anchor=north west][inner sep=0.75pt]  [font=\fontsize{0.78em}{0.94em}\selectfont]   {$e^{\{1\}}_{+l}$};
\draw (76.08,269.34) node [anchor=north west][inner sep=0.75pt]  [font=\fontsize{0.78em}{0.94em}\selectfont]  {$ent^{\{1\}}_{s}$};
\draw (111.17,269.34) node [anchor=north west][inner sep=0.75pt]  [font=\fontsize{0.78em}{0.94em}\selectfont]   {$ent^{\{1\}}_{e}$};
\draw (320.65,274.34) node [anchor=north west][inner sep=0.75pt]   [align=left] {...};
\draw (413.85,269.34) node [anchor=north west][inner sep=0.75pt]  [font=\fontsize{0.78em}{0.94em}\selectfont]   {$s^{\{m\}}_{-l}$};
\draw (448,269.34) node [anchor=north west][inner sep=0.75pt]  [font=\fontsize{0.78em}{0.94em}\selectfont]   {$e^{\{m\}}_{-l}$};
\draw (514.58,269.34) node [anchor=north west][inner sep=0.75pt]  [font=\fontsize{0.78em}{0.94em}\selectfont]   {$s^{\{m\}}_{+l}$};
\draw (549.74,269.34) node [anchor=north west][inner sep=0.75pt]  [font=\fontsize{0.78em}{0.94em}\selectfont]   {$e^{\{m\}}_{+l}$};
\draw (336.57,269.34) node [anchor=north west][inner sep=0.75pt]  [font=\fontsize{0.78em}{0.94em}\selectfont]  {$ent^{\{m\}}_{s}$};
\draw (371.15,269.34) node [anchor=north west][inner sep=0.75pt]  [font=\fontsize{0.78em}{0.94em}\selectfont]  {$ent^{\{m\}}_{e}$};
\draw (223.66,274.34) node [anchor=north west][inner sep=0.75pt]   [align=left] {...};
\draw (485.54,274.34) node [anchor=north west][inner sep=0.75pt]   [align=left] {...};

\end{tikzpicture}
\caption{The sketch of the aggregator architecture in the stacking approach.}
\label{fig:framework}
\end{figure}

Figure~\ref{fig:framework} shows the sketch of the proposed model architecture. We use the entropy of each model's output as an engineered feature to take the base models' uncertainty into account, which seems as an exploitable clue to favor one model over another. To exploit the contextual predictions of base models, we use a window of size $2l+1$ centered at the current token. The window size and the number of layers are investigated in the experiments. 

\section{Experiments}
In this section, we first introduce the evaluation setting including the used datasets and models. Then, the experimental results of ensemble methods are presented and discussed. Finally, the out-of-distribution generalization of the ensemble methods is investigated and compared with that of the popular fine-tuning approach.

\subsection{Evaluation setting}
We use three different evaluation settings to investigate the effectiveness of ensemble methods in different situations:
\begin{itemize}
    \item \textit{Heterogeneous}: Different models are trained on the same dataset,
    \item \textit{Homogeneous}: The same model is trained on different datasets, 
    \item \textit{Hybrid}: Different models are trained on different datasets.
\end{itemize}

 Table~\ref{tab:datasets} shows the datasets used in experiments. All datasets, even those extracted from the same source like Wikipedia, have different features. Therefore, as will be seen later, a trained model on one dataset fails at producing acceptable results on others. In all of these datasets, the paragraphs are concatenated as a single context. The data collection setting is the same as that in the MRQA shared task~\citep{RN230}\footnote{\url{https://github.com/mrqa/MRQA-Shared-Task-2019}}. =In the rest of the paper, the SQuAD 1.1 dataset is referred to as SQuAD for brevity.

\begin{table*}[ht!]
\centering 
\begin{tabular*}{\textwidth}{m{3.4cm}@{\hspace{0.1cm}} m{2.05cm} @{\hspace{0.3cm}} m{2.0cm} @{\hspace{0.3cm}} m{0.9cm} @{\hspace{0.3cm}} m{0.9cm} @{\hspace{0.3cm}} m{0.5cm} @{\hspace{0.15cm}} m{0.9cm}}
\hline \textbf{Datasets} & \textbf{Q Source} &  \textbf{ C Source} & \textbf{Train} & \textbf{Test} & \textbf{\textbar Q\textbar} & \textbf{\textbar C\textbar} \\ \hline
TriviaQA \citep{RN61} & Trivia & Web snippets & 61,688 & 7,785 & 16 & 784 \\
SQuAD 1.1 \citep{RN22} & Crowdsourced & Wikipedia & 86,588 & 10,507 & 11 & 137\\
NewsQA \citep{RN14} & Crowdsourced & News articles & 74,160 & 4,212 & 8 & 599 \\
Natural Questions \citep{RN242} & Search logs & Wikipedia & 104,071 & 12,836 & 9 & 153 \\
DROP \citep{RN206} & Crowdsourced & Wikipedia & 77,409& 1,503& 11 & 243 \\
DuoRC \citep{RN243} & Crowdsourced & Movie plot & 60,721&1,501 & 9 & 681 \\
HotpotQA-gold \citep{RN244} & Crowdsourced & Wikipedia & 72,928 & 5,904 & 22 & 232 \\
HotpotQA-all \citep{RN244} & Crowdsourced & Wikipedia & - & 7,405 & 22 & 1,174\\
RACE \citep{lai2017race} & Experts & Examinations & - & 674 & 12 & 349 \\
TextbookQA \citep{kembhavi2017you} & Experts & Textbook & - & 1,503 & 11 & 657 \\
BioASQ \citep{tsatsaronis2015overview} & Experts & Science & - & 1,504 & 11 & 248 \\
RelationExtraction \citep{levy-etal-2017-zero} & Synthetic & Wikipedia & - & 2,948 & 9 & 30 \\\hline
\end{tabular*}
\caption{\label{dataset} Datasets used in the experiments. HotpotQA-gold contains only two gold paragraphs as the context, while HotpotQA-all contains these two gold paragraphs along with eight distractor paragraphs. The HotpotQA-all dataset as well as the training sets of DROP and DuoRC datasets are downloaded from MultiQA project \cite{talmor-berant-2019-multiqa}, because they are not available in the MRQA shared task data. \textbar Q\textbar\space and \textbar C\textbar\space are the average number of words in questions and contexts, respectively.}
\label{tab:datasets}
\end{table*}

As the base models, a wide variety of popular systems are employed including BiDAF~\citep{RN2}, DrQA~\citep{RN57}, QANet~\citep{RN50}, NAQANet~\citep{RN206}, and two versions of the BERT model, i.e., BERT-large~\citep{devlin2018bert} and DistilBERT~\citep{RN245}. The AllenNLP python library\footnote{\url{https://github.com/allenai/allennlp-reading-comprehension}} is used to train and test BiDAT, QANet, and NAQANet models; while for the DrQA and BERT models, the Facebook Research library\footnote{\url{https://github.com/facebookresearch/DrQA}} and the Hugging Face Transformers library\footnote{\url{https://github.com/huggingface/transformers}} are used, respective ly.

We use two common extractive MRC metrics to evaluate the learned models. These include EM (Exact Match), which is the percentage of samples that are completely correctly answered; and F1, which is the harmonic mean of precision and recall at word level:
\begin{equation}\label{eq:f1}  
    \begin{aligned}
    F1= 2 \times \frac {precision\times recall}{precision + recall},
    \end{aligned}
\end{equation}
where precision is the fraction of tokens extracted by the model that are also present in the ground-truth, and recall is the fraction of all ground-truth tokens that are extracted.

\subsection{Preprocessing}
The first step to ensemble different models is to unify their outputs. As mentioned earlier, in the extractive MRC systems, each model outputs two probability distributions over context tokens as the start and end probabilities. However, different models may use different tokenizers. For example, the AllenNLP NAQANet model trained on DROP dataset splits the numeric expression ``12-135” into three tokens, ``12”, ``-”, and ``135”;
while in DrQA and BiDAF models, the whole expression is considered as one token. 

To address this issue, we perform ensemble techniques in character level by assigning the probability value of every
token to each of its constituting characters.
We also ignore unnecessary characters like blank as well as those that are only available in some models' outputs. The special
characters, like non-English ones, which are converted to specific symbols in some models are also removed in the preprocessing phase.

\subsection{Results and discussion}
As stated earlier, the experiments are performed in three different settings. In the \textit{heterogeneous} setting, we train five different models including DrQA, BiDAF, QANet, BERT-large, and DistilBERT on the SQuAD dataset.
In the \textit{homogeneous} setting, we train the simple and popular BiDAF model on five different datasets including SQuAD, NewsQA, Natural Questions, DROP, and DuoRC. Finally, in the \textit{hybrid} setting, we train  BiDAF on NewsQA, DrQA on SQuAD, NAQANet on DROP, QANet on Natural Questions, and distilBERT on TriviaQA. To refer to a dataset or model, we use the first two letters of one-word names, and first letters of the first two words for multi-words names. For example, "QANet" and "Natural Questions" are abbreviated as QA and NQ, respectively. 

All the trained models and ensemble methods are tested on datasets from a different domain to evaluate their generalization. The test datasets are HotpotQA-gold, HotpotQA-all, and TriviaQA in the homogeneous setting; HotpotQA-gold, HotpotQA-all, and NewsQA in the heterogeneous setting; and HotpotQA-gold, HotpotQA-all, and RACE in the hybrid setting.

\subsubsection{Different combinations of base models}
We start the experiments with investigating different combinations of base models. As the ensemble method, the simple multiplication operator is used (Eq.~\ref{eq:simple}), while different ensemble techniques are investigated in Section~\ref{subsec:dif-ensemble}. For brevity in this section, some representative combinations are presented in Table~\ref{tab:g1}, while Table~\ref{tab:A-g1} in Appendix shows more combinations.
Note that because of different based models used in different evaluation settings, their results are not comparable. Instead, the ensemble methods in each setting should be compared with its own base models. 

In Table~\ref{tab:g1-heterogeneous}, we have a pool of very diverse base models in terms of accuracy. For example, based on the EM measure, the BERT model trained on the SQuAD dataset (BERT-SQuAD in Table~\ref{tab:g1-heterogeneous}) is about 15\% more accurate than the second best model. Therefore, ensemble of any subset of models that contains BERT-SQuAD (BE) cannot obtain a better result than the best model. However, in cases that the accuracies of base models are close, the ensemble has a positive impact. For example, the BiDAF-NewsQA and BiDAF-DuoRC models have similar accuracies in Table~\ref{tab:g1-homogeneous} which results in more than 3\% increase in EM measure  using the simple multiply-based ensemble (Ne-Du). 

It is worth noting that in addition to the high training cost of the BERT model, it also suffers from a high inference computational cost. For example, our estimation based on 100 instances of HotpotQA development set using an Intel Core i5-6500 CPU shows that the average prediction time of BERT model is 1.11 seconds per sample, while the BiDAF model do this prediction about 10 times faster (0.10 seconds per sample). Since we aimed to use lighter and greener models, we experimented on BERT just to investigate the situations that there are significant gaps between the base models in terms of accuracy.
\begin{table}[ht!]
\small
\centering
\begin{subtable}[b]{\textwidth}
\centering
\begin{tabular}[b]{m{3.5cm} I I I I}
\hline  \multirow{2}{*}{\diagbox{\textbf{Model}}{\textbf{Test Set}}} & \multicolumn{2}{G}{\textbf {In-Domain}}& \multicolumn{2}{G}{\textbf {Out-of-Domain}}
\\\cline{2-3} \cline{4-5} 
  & \textbf{EM} & \textbf{F1} & \textbf{EM} & \textbf{F1} \\
\hline

DrQA-SQuAD&69.4&78.9&24.32&35.85\\
BiDAF-SQuAD&68.35&77.83&31.18&44.30\\
QANet-SQuAD&72.42&81.31&32.06&46.26\\
DistilBERT-SQuAD&79.08&86.85&35.86&54.45\\
BERT-SQuAD&86.92&93.15&51.08&68.82\\
\hline
Dr-QA-DB-BE&83.53&90.40&43.74&60.19\\
Dr-Bi-QA-DB&78.32&86.12&37.67&52.59\\\hline
Dr-Bi-QA&74.15&82.78&32.96&46.66\\
QA-DB-BE&84.54&91.20&44.84&61.93\\\hline
Bi-QA&73.76&82.39&34.37&48.71\\
QA-BE&85.04&91.57&46.82&63.13\\\hline
All&82.88&89.83&41.89&57.53\\\hline
\end{tabular}
\caption{Heterogeneous}
\label{tab:g1-heterogeneous}
\end{subtable}
\newline
\newline
\centering
\small
\begin{subtable}[b]{\textwidth}
\centering
\begin{tabular}[b]{m{3.5cm} I I I I}
\hline  \multirow{2}{*}{\diagbox{\textbf{Model}}{\textbf{Test Set}}} & \multicolumn{2}{G}{\textbf {In-Domain}}& \multicolumn{2}{G}{\textbf {Out-of-Domain}}
\\\cline{2-3} \cline{4-5} 
  & \textbf{EM} & \textbf{F1} & \textbf{EM} & \textbf{F1} \\
\hline
 BiDAF-SQuAD&68.35&77.83& 31.18 &44.30 \\
 BiDAF-NewsQA&40.43&55.53 & 22.55 & 37.31 \\
 BiDAF-Natural Questions&54.60&66.10 & 11.34 & 20.10  \\
 BiDAF-DROP &46.99&52.07& 05.49 & 12.28 \\
 BiDAF-DuoRC &57.87&67.83& 20.20 & 29.63 \\
 \hline
Ne-NQ-DR-Du&-&-&24.47&37.54\\
SQ-Ne-DR-Du&-&-&30.01&43.72\\\hline
SQ-Ne-Du&-&-&28.32&42.38\\
Ne-NQ-Du&-&-&21.96&34.82\\\hline
SQ-Ne&-&- & 30.86 & 45.86 \\
Ne-Du &-&-& 25.62 & 40.46 \\\hline
All&-&-&26.27&39.04\\\hline
\end{tabular}
\caption{Homogeneous}
\label{tab:g1-homogeneous}
\end{subtable}
\end{table}
\clearpage

\begin{table}[ht!]\ContinuedFloat
\small
\centering
\begin{subtable}[b]{\textwidth}
\centering
\begin{tabular}[b]{m{3.5cm} I I I I}
\hline  \multirow{2}{*}{\diagbox{\textbf{Model}}{\textbf{Test Set}}} & \multicolumn{2}{G}{\textbf {In-domain}}& \multicolumn{2}{G}{\textbf {Out-of-Domain}}
\\\cline{2-3} \cline{4-5} 
  & \textbf{EM} & \textbf{F1} & \textbf{EM} & \textbf{F1} \\
\hline
BiDAF-NewsQA&40.43&55.53&22.55&37.31\\
DrQA-SQuAD&69.40&78.90&24.32&35.85\\
NAQANet-DROP&75.05&81.45&12.25&20.79\\
QANet-Natural Questions&56.54&68.84&18.95&31.01\\
distilBERT-TriviaQA&30.25&36.72&20.98&32.23\\\hline
Bi-Dr-QA-D6B&-&-&32.86&47.10\\
Bi-Dr-Na-DB&-&-&30.16&44.42\\\hline
Bi-Dr-DB&-&-&29.08&43.16\\
Dr-QA-DB&-&-&29.93&43.70\\\hline
Bi-Dr&-&-&28.66&42.78\\
Dr-DB&-&-&28.99&44.01\\\hline
All&-&-&30.74&44.55\\\hline
\end{tabular}
\caption{Hybrid}
\label{tab:g1-hybrid}
\end{subtable}
\caption{Investigating different combinations of base models in different evaluation settings. The used ensemble method is the simple multiplication (Eq.~\ref{eq:simple}).In all tables, the HotpotQA-gold dataset is used as the out-of-domain test set. The hyphen symbol (-) means not-applicable in-domain evaluation due to different source domains used in the respective base models. An extended version of this table is presented in Table~\ref{tab:A-g1}.}
\label{tab:g1}
\end{table}

The results obtained in the hybrid setting (Table~\ref{tab:g1-hybrid}) are different from the previous two ones. Here, the base models are in a closer range of accuracies which results in significant improvements in all cases. For example, the ensemble of all base models obtains 7.24\% absolute F1 improvement compared to the best base model. Also, if we exclude the weakest base model (NAQANet-DROP), this improvement increases to about 10\%.

\begin{figure}[!ht]
\centering
\vspace{-0.2cm}
\includegraphics[width=\linewidth,valign=T]{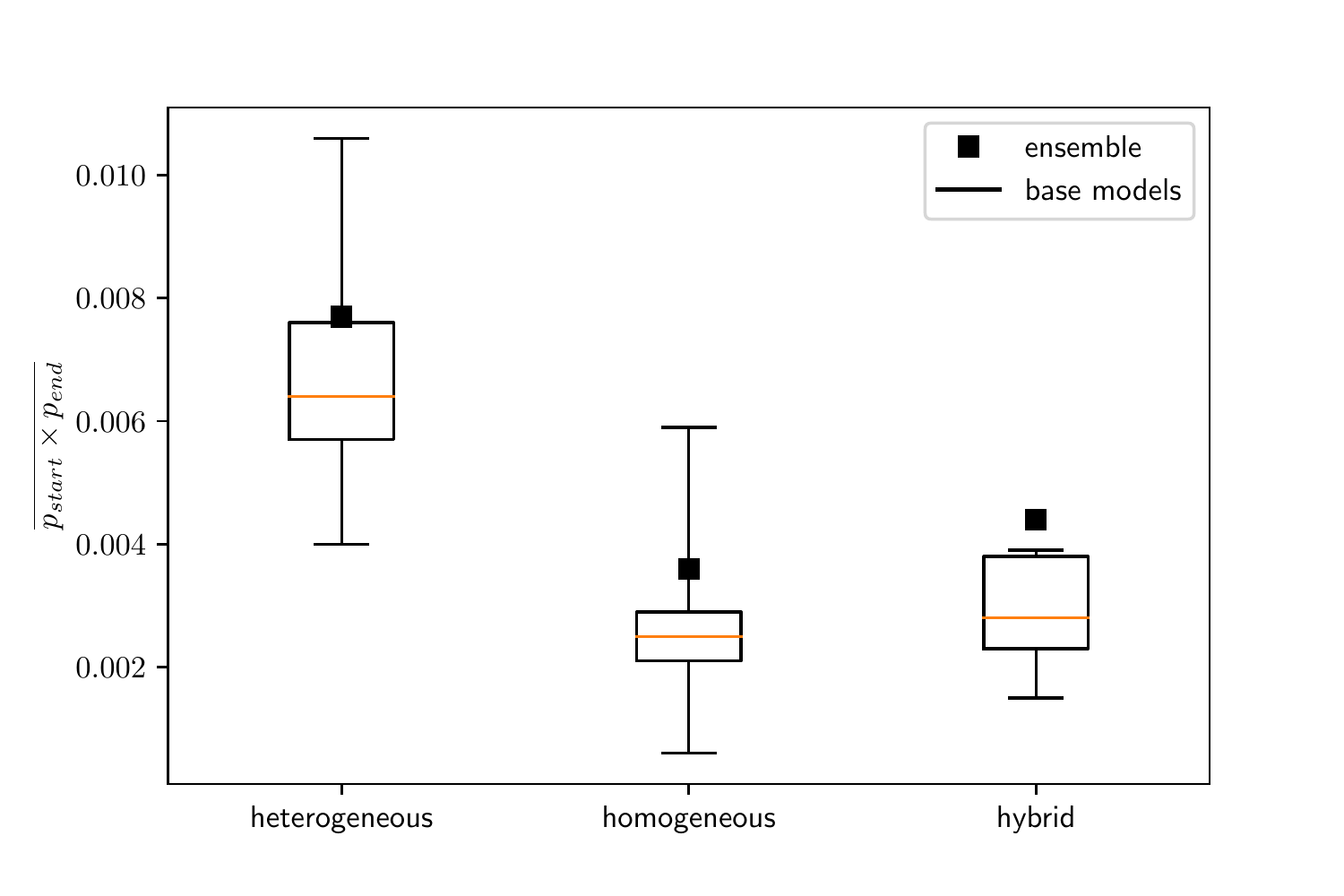} 
\caption{The average over the ground-truth span of $p_{start}\times p_{end}$ generated by base models as well as the multiply-based ensemble in different evaluation settings. The mean value for the ensemble method is shown with square mark, while the ranges of mean values for different base models are illustrated with boxplots.}
\label{fig:multiply_analysis}
\end{figure}

To better explain the reason behind the difference between the performance of the multiply-based ensemble method in different settings, we investigate the probability distributions produced by the models in the ground-truth span. Figure~\ref{fig:multiply_analysis} shows the average over the ground-truth span of $p_{start}\times p_{end}$ generated by base models as well as the multiply-based ensemble in different evaluation settings. The mean value for the ensemble method is shown with square mark, while the ranges of mean values for different base models are illustrated with vertical bars. Note that to make the outputs of base models and the ensemble method comparable, we use \textit{geometric mean} instead of pure multiplication as the ensemble function. Even though the values generated by geometric mean and multiplication are different, they give the same final decisions regarding the most probable spans.

As shown in this figure, in the hybrid setting, the ensemble method obtains higher average value of $p_{start}\times p_{end}$ over the ground-truth span than the base models, which is inline with its better performance in the hybrid setting compared to the other two ones (Tables~\ref{tab:g1}). The reason for worse performance of the ensemble method in the heterogeneous and homogeneous settings is that $p_{start}\times p_{end}$ values obtained by base models in these settings are spread over wider ranges, which prevents the ensemble method from improving the best base model. In fact, a small value produced by one of the base models can have a huge negative impact on the product in the multiply-based ensemble method. That is why, as shown in Tables ~\ref{tab:g1-homogeneous}~and~\ref{tab:g1-hybrid}, even though the best base model in the homogeneous setting has 7\% higher F1 than the best base model in the hybrid setting, the ensemble model in the hybrid setting beats the ensemble and base model in the homogeneous setting.

As the final investigation in this section, we estimated the maximum achievable improvements using ensemble models. For this, we chose the most accurate base model for each test sample exploiting its ground-truth answer. The EM of this system is 63.51, 49.45, and 50.02 in heterogeneous, homogeneous, and hybrid settings. The respective F1 values are also 77.22, 65.90, and 65.43.
Even though the ground-truth answers are not accessible in real-world situations, the high accuracy of this hypothetical system shows the potential capability of the ensemble approach to make high improvements in generalization for unseen datasets.

\subsubsection{Different ensemble approaches}
\label{subsec:dif-ensemble}
\begin{figure}
\centering
\begin{subfigure}[t]{.32\linewidth}
\includegraphics[width=\linewidth,valign=T]{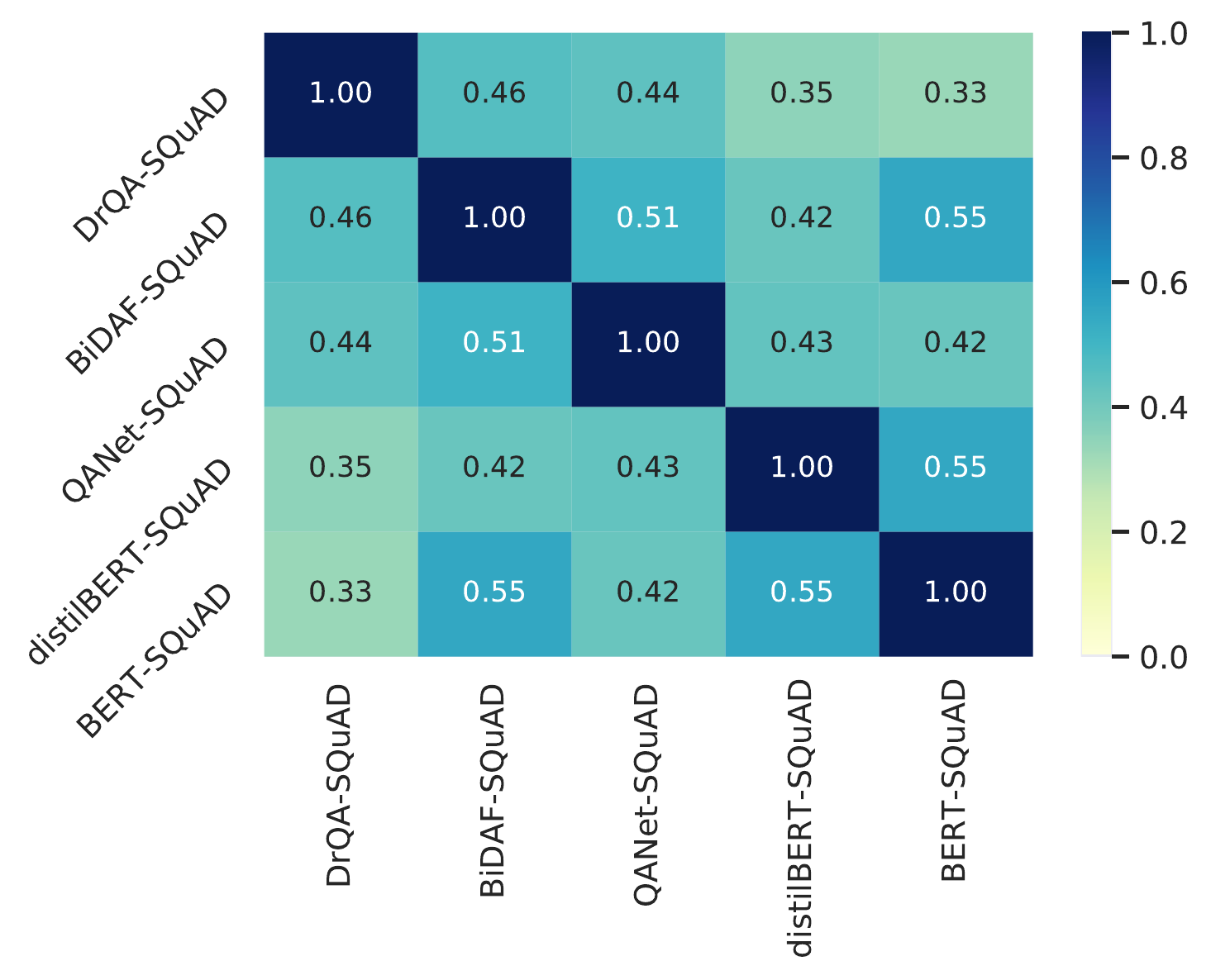} 
\hfill
\caption{Heterogeneous}
\label{fig:jaccard-heterogeneous}
\end{subfigure}
\begin{subfigure}[t]{.32\linewidth}
\includegraphics[width=\linewidth,valign=T]{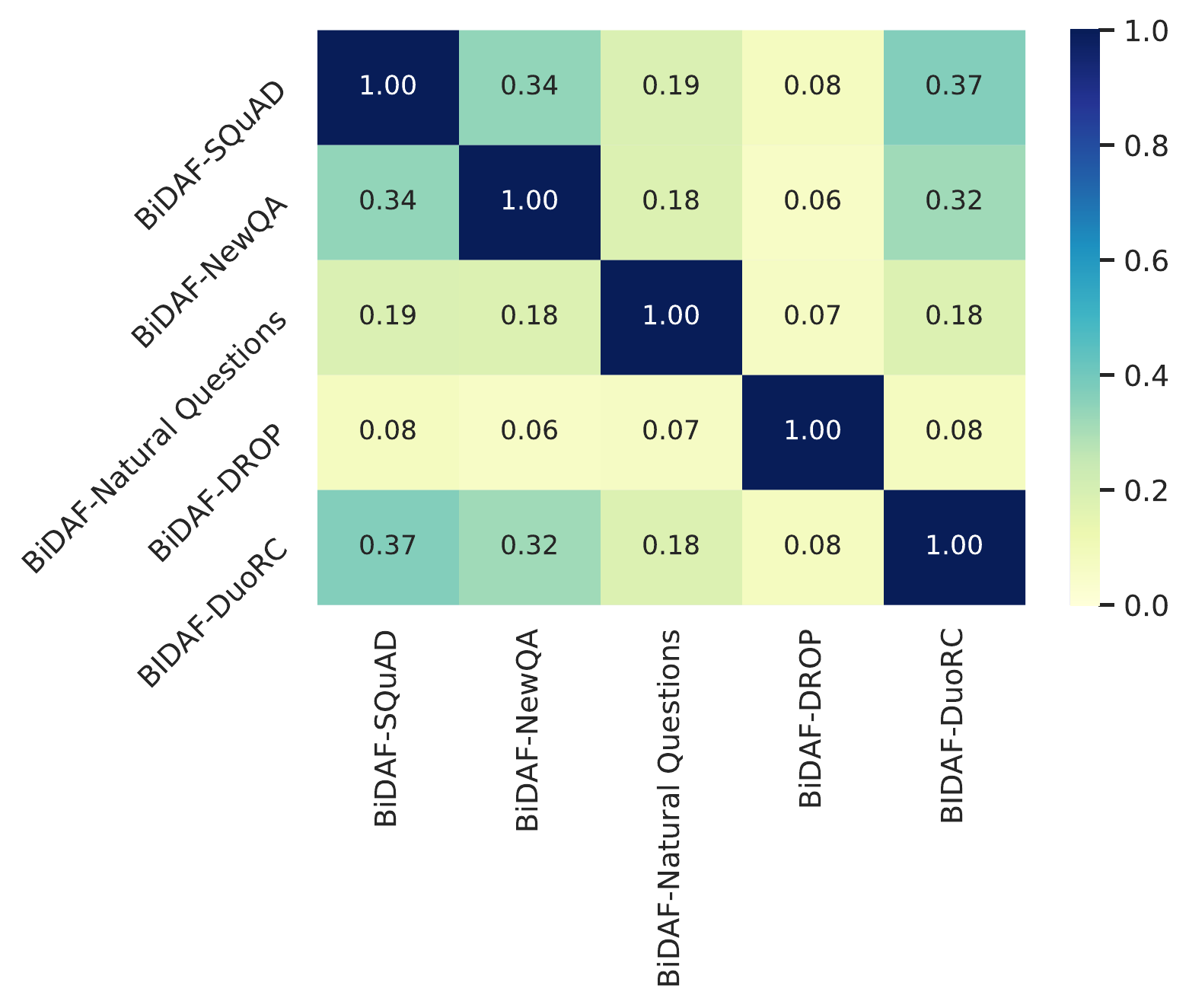} 
\caption{Homogeneous}
\label{fig:jaccard-homogeneous}
\end{subfigure}
\begin{subfigure}[t]{.32\linewidth}
\includegraphics[width=\linewidth,valign=T]{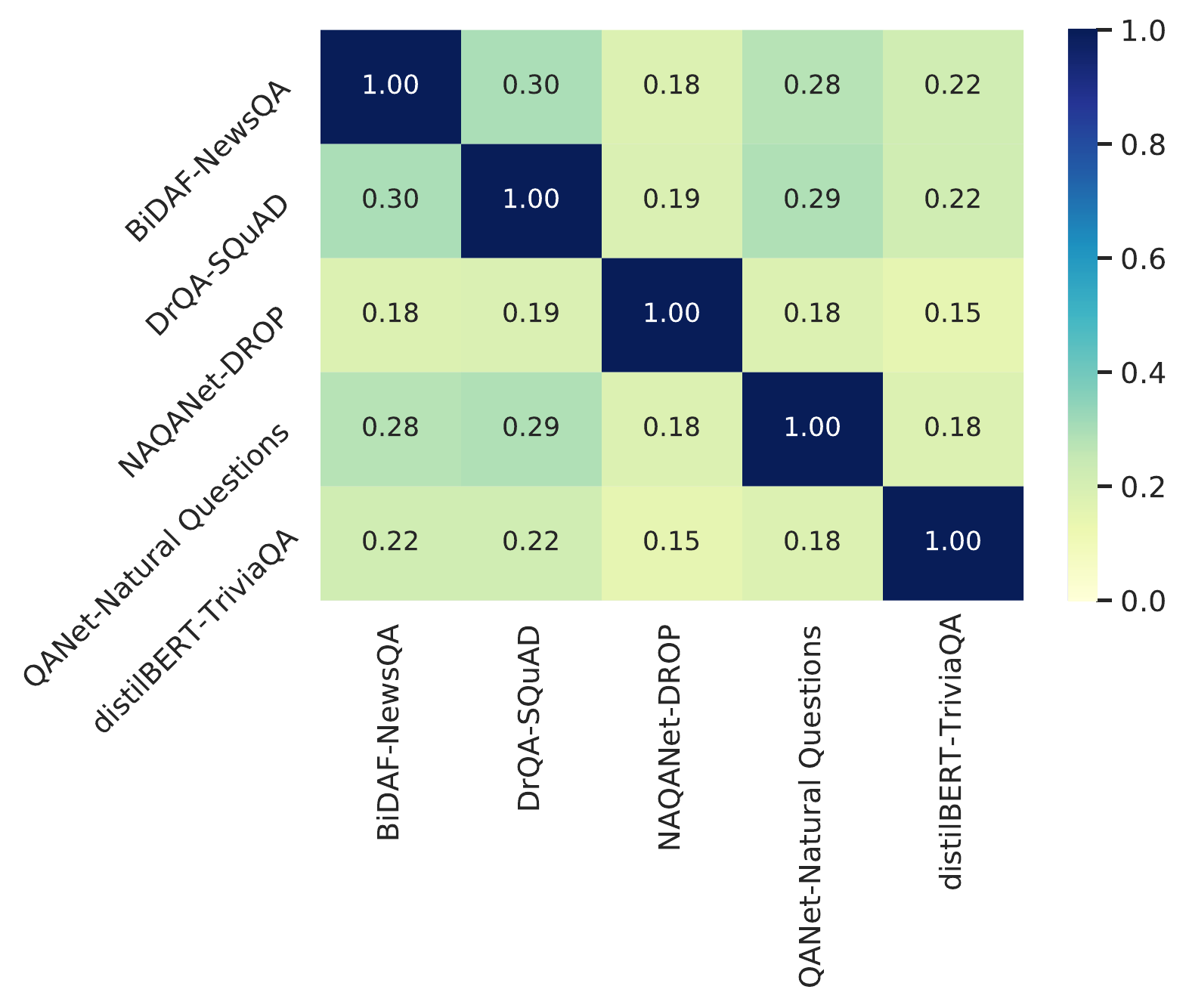} 
\caption{Hybrid}
\label{fig:jaccard-hybrid}
\end{subfigure}
\caption{The Jaccard similarity between pairs of base models in different evaluation settings. The average of non-diagonal elements is 0.446, 0.187, and 0.219 for heterogeneous, homogeneous, and hybrid settings, respectively.}
\label{fig:jaccard}
\end{figure}

In this section, we present and discuss the results of applying different ensemble approaches in heterogeneous, homogeneous, and hybrid settings. First, to obtain an insight into the base models in different settings, Figure~\ref{fig:jaccard} illustrates the \textit{Jaccard} similarity between the results of each pair of base models, computed as the ratio of the number of test samples correctly answered by both models to the number of test samples correctly answered by either of the models. We used HotpotQA-gold as the test set in this experiment. According to this figure, base models are significantly more similar in the heterogeneous setting than in the other two settings. This shows that difference in training datasets of base models (as in homogeneous and hybrid settings) plays a more effective role in creating a diverse pool than difference in the model structures.

The results of applying different ensemble approaches for heterogeneous, homogeneous, and hybrid settings are presented in Table~\ref{tab:g3}, respectively. First, we compare the four element-wise functions for probabilistic equal-weight ensemble, i.e., \textit{mean}, \textit{multiply}, \textit{max}, and \textit{min} (Eq.~\ref{eq:simple}). These are the methods prefixed with \textit{prob-equal-weight} in the tables. The probabilistic approach means that the start and end probabilities generated by base models are directly used in the ensemble functions.
In the heterogeneous setting (Table~\ref{tab:g3-heterogeneous}), the results of different aggregators are not so different; while in the homogeneous (Table~\ref{tab:g3-homogeneous}) and the hybrid (Table~\ref{tab:g3-hybrid}) settings, we can observe a more diverse performance. The two best aggregators among these four functions are \textit{multiply} and \textit{mean}, where the \textit{multiply} function obtains the best results in the hybrid setting even in comparison with sophisticated weighting techniques.

To examine the non-probabilistic approach, we consider the one-hot coding of the base models' outputs as the input to the ensemble component. That is, for each of the probability distributions generated by base models, the largest element is set to 1, and the other elements are set to zero. We just report the results obtained by the \textit{mean} function (non-prob-equal-weight-mean) since the other functions are highly affected by zeros. According to the tables, as expected, the probabilistic approach is superior in nearly all cases.

For the unequal weighting and stacking methods that we need a subset of samples from the new distribution, we use a subset of 10k samples from the train set of the target distribution, and evaluate the models on the target test set. In the unequal weighting approach, the train set is used to estimate the accuracy of the base models which is exploited in Eq.~\eqref{eq:weighting}. In the stacking approach, this set is employed to train a multi-layer feed-forward neural network whose sketch is depicted in Figure~\ref{fig:framework}. Figure~\ref{fig:A-stackingStructures} shows the results of evaluating the number of layers and the window size in this architecture for a representative setting. The Adam optimizer~\citep{ADAM} with learning rate 0.0001 is used in the learning process. 

Two unequal weighting methods are evaluated in Table~\ref{tab:g3}. In \textit{prob-unequal-weight-fixed}, a similar approach to \citep{RN261} is followed, where a fixed value is used as $\alpha$ in Eq.~\eqref{eq:weighting}, and their averages performance is reported.
 In \textit{prob-unequal-weight-auto} method on the other hand, the automatic tuning presented in Eq.~\eqref{eq:auto} is exploited to determine the value of $\alpha$. According to these tables, the proposed automatic tuning method is superior to the fixed one in most cases.
 
 To sum this section up, the results presented in Table~\ref{tab:g3} show that where there is not a base model significantly more accurate than others, ensemble learning is an effective approach in enhancing the generalization of MRC systems to new domains. In the heterogeneous setting (Tables~\ref{tab:g3-heterogeneous}), BERT-SQuAD is far more accurate than other base models, so the ensemble is destructive. In the homogeneous setting (Tables~\ref{tab:g3-homogeneous}), the proposed stacking method is superior to the base models and other ensemble approaches. Finally, in the hybrid setting (Table~\ref{tab:g3-hybrid}), the simple multiply-based ensemble obtains the best results in almost all cases, while the stacking method is comparable to it.

\subsubsection{Out-of-distribution generalization robustness}
In this section, we investigate the robustness of the ensemble approach to the change in the target distribution compared to the popular fine-tuning approach. To this end, we develop unequal weighting ensemble models in the homogeneous setting, using 5k samples from the out-of-distribution labeled dataset to determine models' weights. Then, a different dataset than the base models' and ensemble's training datasets is used as the test set to measure out-of-distribution generalization. As the competing approach, the best base model is fine-tuned and tested with the same datasets. The results, as shown in Table~\ref{tab-group2}, indicate that the ensemble method is more robust against data distribution change than all base models and the fine-tuning approach.

\begin{table}[!htbp]
\footnotesize
\begin{subtable}[b]{\textwidth}
\centering
\begin{tabular}{@{}l*{2}{r@{.}l}{c}*{2}{r@{.}l}{c}*{2}{r@{.}l}@{}}
\toprule
\multirow{2}{*}{\diagbox{\textbf{Model}}{\textbf{Test Set}}} & \multicolumn{4}{c}{HotpotQA-all} && \multicolumn{4}{c}{HotpotQA-gold} && \multicolumn{4}{c}{NewsQA}\tabularnewline
\cmidrule{2-5} \cmidrule{7-10} \cmidrule{12-15}
 & \multicolumn{2}{c}{EM} & \multicolumn{2}{c}{F1} && \multicolumn{2}{c}{EM} & \multicolumn{2}{c}{F1} && \multicolumn{2}{c}{EM} & \multicolumn{2}{c}{F1}\tabularnewline
\midrule  
DrQA-SQuAD & 15&41  & 24&18  &&  24&32                & 35&85&&  29&83               & 44&14 \tabularnewline
BiDAF-SQuAD & 18&61 & 27&53 &&  31&18              & 44&30&&  27&69 & 42&36 \tabularnewline
QANet-SQuAD & 19&95 & 29&99 &&  32&06 & 46&26 && 30&59&45&38   \tabularnewline
distilBERT-SQuAD & 21&39 & 33&57 &&  35&86 & 54&45 && 33&59&49&20    \tabularnewline
BERT-SQuAD & 30&68 & \textbf{44}&\textbf{43}&& 51&08&\textbf{68}&\textbf{82} && 40&29&   \textbf{57}&\textbf{35} \tabularnewline
\bottomrule

prob-equal-weight-mean&25&62&37&04&&40&89&56&20&&37&34&52&53 \tabularnewline
prob-equal-weight-mul&26&58&38&41&&41&89&57&53&&37&89&53&23 \tabularnewline
prob-equal-weight-max&25&36&36&75&&40&43&55&61&&36&70&51&81 \tabularnewline
prob-equal-weight-min&25&27&37&29&&41&77&57&72&&35&89&51&51 \tabularnewline
non-prob-equal-weight-mean&24&67&36&83&&38&67&53&80&&35&59&50&75 \tabularnewline
\bottomrule
prob-unequal-weight-fixed&\textbf{31}&\textbf{06}&43&77&&51&63&68&15&&\textbf{40}&\textbf{77}&56&55 \tabularnewline
prob-unequal-weight-auto&30&47&43&01&&51&63&68&15&&38&95&54&61 \tabularnewline
prob-stacking&30&82&43&50&&50&06&67&34&&39&71&55&68 \tabularnewline

non-prob-unequal-weight-auto&30&41&43&52&&\textbf{52}&\textbf{47}&68&65&&38&70&54&31 \tabularnewline
\bottomrule
\end{tabular}
\caption{Heterogeneous}
\label{tab:g3-heterogeneous}
\end{subtable}
\newline
\newline
\newline
\begin{subtable}[b]{\textwidth}
\centering
\begin{tabular}{@{}l*{2}{r@{.}l}{c}*{2}{r@{.}l}{c}*{2}{r@{.}l}@{}}
\toprule
\multirow{2}{*}{\diagbox{\textbf{Model}}{\textbf{Test Set}}} & \multicolumn{4}{c}{HotpotQA-all} && \multicolumn{4}{c}{HotpotQA-gold} && \multicolumn{4}{c}{TriviaQA}\tabularnewline
\cmidrule{2-5} \cmidrule{7-10} \cmidrule{12-15}
 & \multicolumn{2}{c}{EM} & \multicolumn{2}{c}{F1} && \multicolumn{2}{c}{EM} & \multicolumn{2}{c}{F1} && \multicolumn{2}{c}{EM} & \multicolumn{2}{c}{F1}\tabularnewline
\midrule  
BiDAF-SQuAD&18&61&27&53&&31&18&44&30&&31&29&40&91 \tabularnewline
BiDAF-NewQA&10&91&19&68&&22&55&37&31&&21&14&31&79 \tabularnewline
BiDAF-Natural Questions&7&28&13&09&&11&34&20&10&&12&19&18&03 \tabularnewline
BiDAF-DROP&05&50&09&85&&05&49&12&28&&06&92&12&55 \tabularnewline
BIDAF-DuoRC&10&79&17&01&&20&20&29&63&&22&04&28&72 \tabularnewline

\bottomrule

prob-equal-weight-mean&16&53&24&85&&29&59&42&41&&32&63&42&68 \tabularnewline
prob-equal-weight-mul&15&23&24&35&&26&27&39&04&&31&43&42&47 \tabularnewline
prob-equal-weight-max&15&43&23&64&&23&66&35&05&&26&34&35&94 \tabularnewline
prob-equal-weight-min&10&21&19&99&&18&08&32&90&&20&63&32&52 \tabularnewline
non-prob-equal-weight-mean&16&39&25&77&&25&44&37&86&&26&60&36&64 \tabularnewline
\bottomrule
prob-unequal-weight-fixed&19&73&29&05&&32&47&45&95&&33&05&43&05 \tabularnewline
prob-unequal-weight-auto&20&05&29&38&&32&42&46&10&&33&24&43&12 \tabularnewline
prob-stacking&\textbf{20}&\textbf{77}&\textbf{30}&\textbf{40}&&\textbf{32}&\textbf{79}&\textbf{46}&\textbf{59}&&\textbf{33}&\textbf{91}&\textbf{43}&\textbf{38}  \tabularnewline
non-prob-unequal-weight-auto&19&12&28&26&&31&65&45&12&&31&63&41&63 \tabularnewline
\bottomrule
\end{tabular}
\caption{Homogeneous}
\label{tab:g3-homogeneous}
\end{subtable}
\end{table}
\pagebreak
\begin{table}[ht!]\ContinuedFloat
\footnotesize
\begin{subtable}[b]{\textwidth}
\centering
\begin{tabular}{@{}l*{2}{r@{.}l}{c}*{2}{r@{.}l}{c}*{2}{r@{}l}@{}}
\toprule
\multirow{2}{*}{\diagbox{\textbf{Model}}{\textbf{Test Set}}} & \multicolumn{4}{c}{HotpotQA-all} && \multicolumn{4}{c}{HotpotQA-gold} && \multicolumn{4}{c}{RACE}\tabularnewline
\cmidrule{2-5} \cmidrule{7-10} \cmidrule{12-15}
 & \multicolumn{2}{c}{EM} & \multicolumn{2}{c}{F1} && \multicolumn{2}{c}{EM} & \multicolumn{2}{c}{F1} && \multicolumn{2}{c}{EM} & \multicolumn{2}{c}{F1}\tabularnewline
\midrule  
BiDAF-NewsQA&10&91&19&68&&22&55&37&31&&8.&46&21.&75 \tabularnewline
DrQA-SQuAD&15&41&24&18&&24&32&35&85&&17.&95&29.&14 \tabularnewline
NAQANet-DROP&7&55&12&55&&12&25&20&79&&05.&93&11.&30 \tabularnewline
QANet-Natural Questions&10&22&18&15&&18&95&31&01&&09.&20&17.&79 \tabularnewline
distilBERT-TriviaQA &11&19&18&16&&20&98&32&23&&10.&53&17.&91 \tabularnewline
\bottomrule
prob-equal-weight-mean&17&41&26&99&&27&81&41&06&&17.&06&27.&16 \tabularnewline
prob-equal-weight-mul&\textbf{19}&\textbf{73}&\textbf{29}&\textbf{96}&&\textbf{31}&\textbf{33}&45&30&&\textbf{18.}&\textbf{99}&\textbf{30.}&\textbf{08}\tabularnewline
prob-equal-weight-max&15&95&25&55&&25&91&39&52&&14.&98&25.&28 \tabularnewline
prob-equal-weight-min&13&73&24&58&&24&54&40&03&&13.&50&22.&64 \tabularnewline
non-prob-equal-weight-mean&14&99&24&38&&28&32&41&87&&14.&39&25.&01 \tabularnewline
\bottomrule
prob-unequal-weight-fixed&15&92&25&47&&28&08&41&75&&-& &-
 \tabularnewline
prob-unequal-weight-auto&16&50&26&00&&28&89&42&75&&-& &- \tabularnewline
prob-stacking&18&70&28&70&&31&04&\textbf{45}&\textbf{31}&&-& &- \tabularnewline
non-prob-unequal-weight-auto&15&48&24&72&&27&83&41&29&&-& &- \tabularnewline
\bottomrule
\end{tabular}
\caption{Hybrid}
\label{tab:g3-hybrid}
\end{subtable}
\caption{The results of applying different ensemble approaches in different evaluation settings. In the hybrid setting, the last four methods that require target train data are not applied on the RACE dataset, because it does not have a train set in the MRQA shared task setting~\citep{RN230}.}
\label{tab:g3}
\end{table}

\begin{table}[!ht]
\begin{center}
\begin{tabular}{l c c c}
\hline
\diagbox{\textbf{Model}}{\textbf{Test Set}}&HotpotQA-gold &SearchQA&TriviaQA\\
\hline
BiDAF-SQuAD&44.30&19.27&40.94\\
BiDAF-NewsQA&37.31&14.69&31.79\\
BiDAF-Natural Questions&20.10&13.35&18.04\\
BiDAF-DROP&12.28&2.56&12.55\\
BiDAF-DuoRC&29.63&14.43&28.72\\
\hline
fine-tune-HotpotQA-gold&-&18.15&35.49\\
fine-tune-SearchQA&36.42&-&36.99\\
fine-tune-TriviaQA&36.43&21.63&-\\
\hline
ensemble-weighting-HotpotQA-gold&-&\textbf{21.97}&\textbf{43.40}\\
ensemble-weighting-SearchQA&45.81&-&43.26\\
ensemble-weighting-TriviaQA&\textbf{46.23}&19.95&-\\
\hline
\end{tabular}
\end{center}
\caption{F1 score of base models, fine-tuning approach, and unequal weighting ensemble method. The fine-tuning and weight calculation are performed using HotpotQA-gold, SearchQA, and TriviaQA datasets; while the test is done on completely unseen datasets.}
\label{tab-group2}
\end{table}

As another experiment, we examine the generalization robustness of models trained on a combination of different datasets with 5k samples from each. Three datasets including HotpotQA-gold, SearchQA, and TriviaQA are used to calculate weights for the unequal weighting ensemble method as well as to fine-tune the best base model. These models are then tested on four new datasets including RACE, TextbookQA, BioASQ, and RelationExtraction. The results are shown in Table~\ref{tab-group3}. According to this table, combination of different datasets does not change the superiority of the ensemble method over fine-tuning compared to using single datasets (Table~\ref{tab-group2}).

\begin{figure}[!htbp]
\centering
\begin{subfigure}[b]{.45\linewidth}
\includegraphics[width=\linewidth,valign=T]{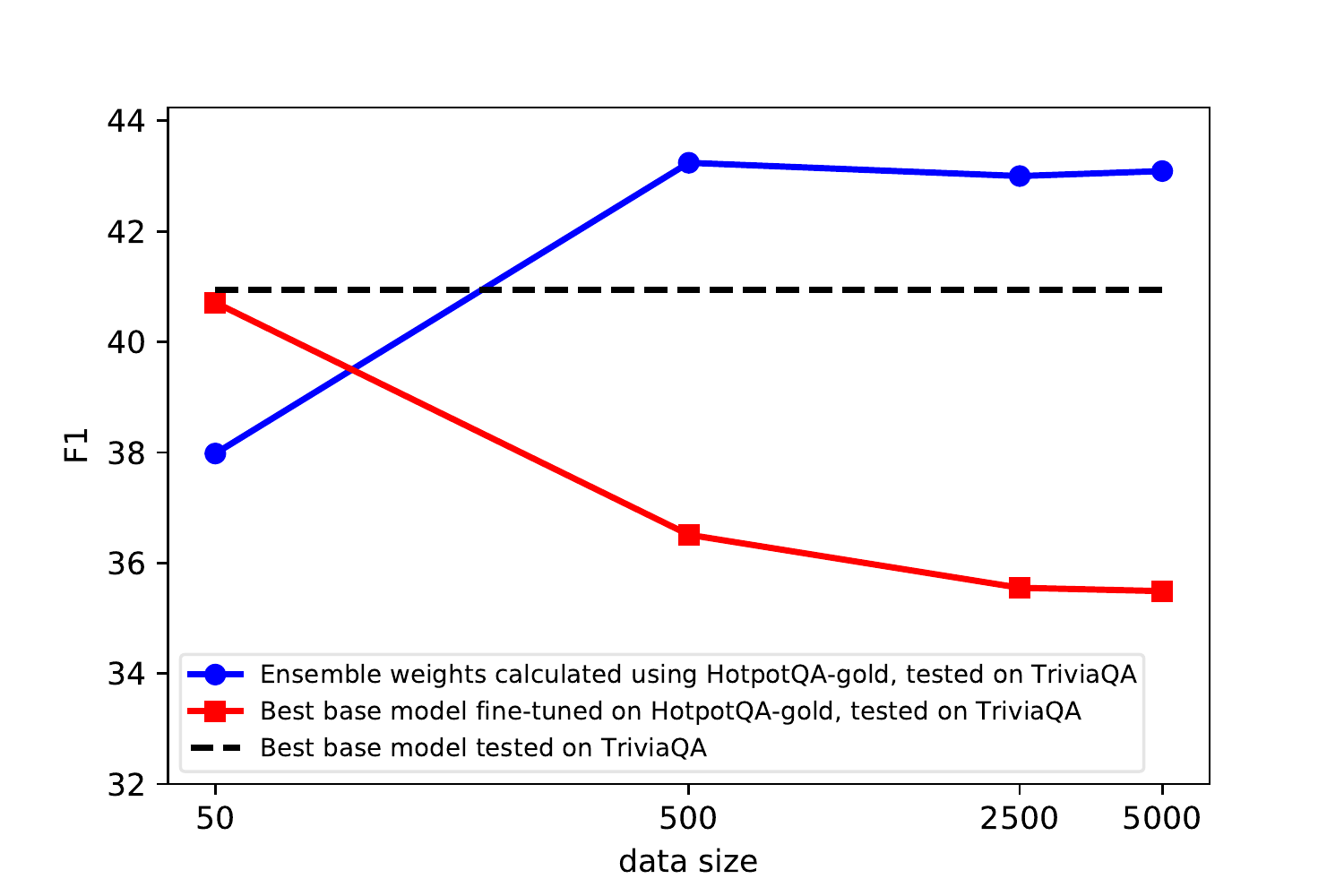} 
\end{subfigure}
\begin{subfigure}[b]{.45\linewidth}
\includegraphics[width=\linewidth,valign=T]{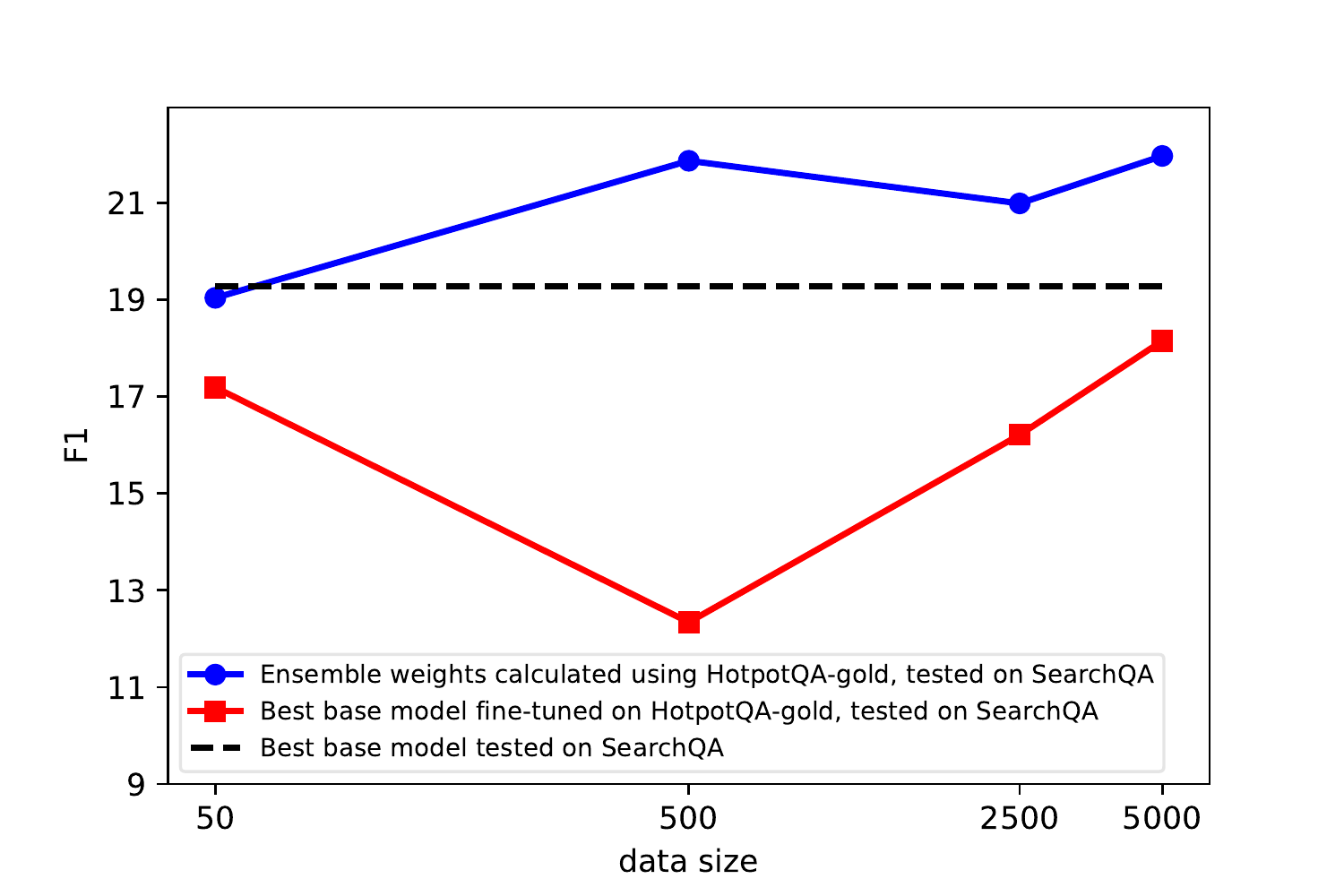} 
\end{subfigure}
\begin{subfigure}[b]{.45\linewidth}
\includegraphics[width=\linewidth,valign=T]{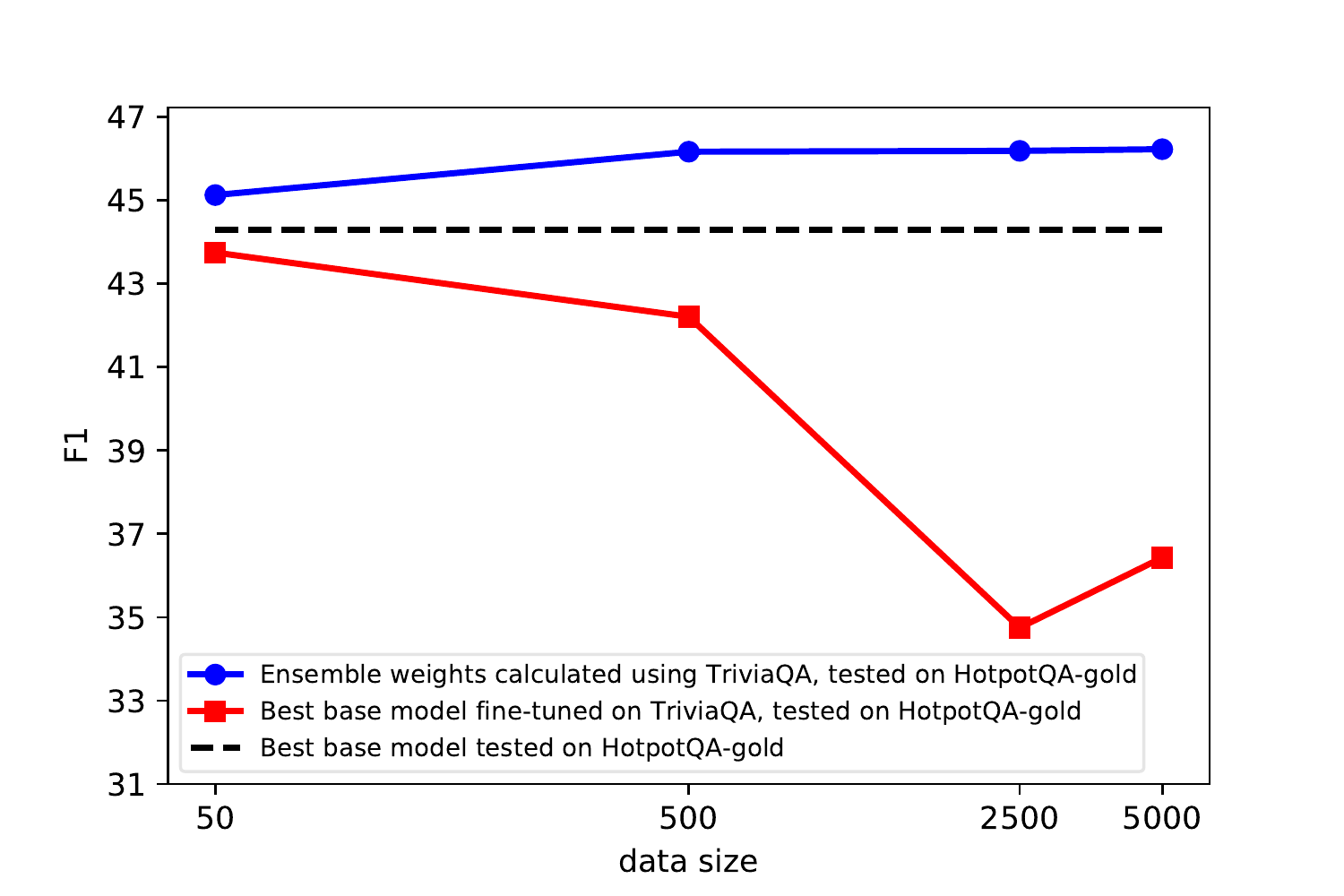} 
\end{subfigure}
\begin{subfigure}[b]{.45\linewidth}
\includegraphics[width=\linewidth,valign=T]{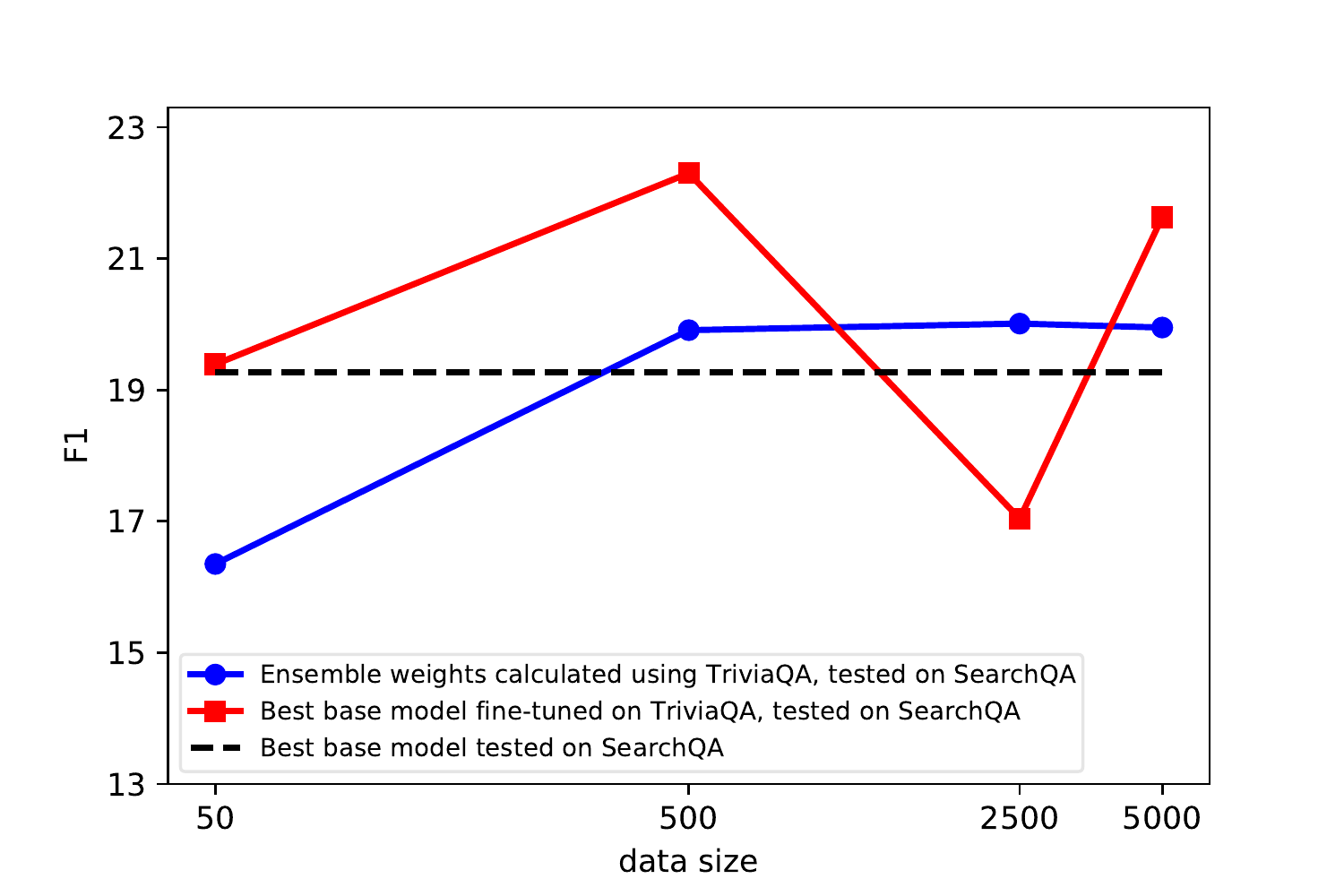} 
\end{subfigure}
\begin{subfigure}[b]{.45\linewidth}
\includegraphics[width=\linewidth,valign=T]{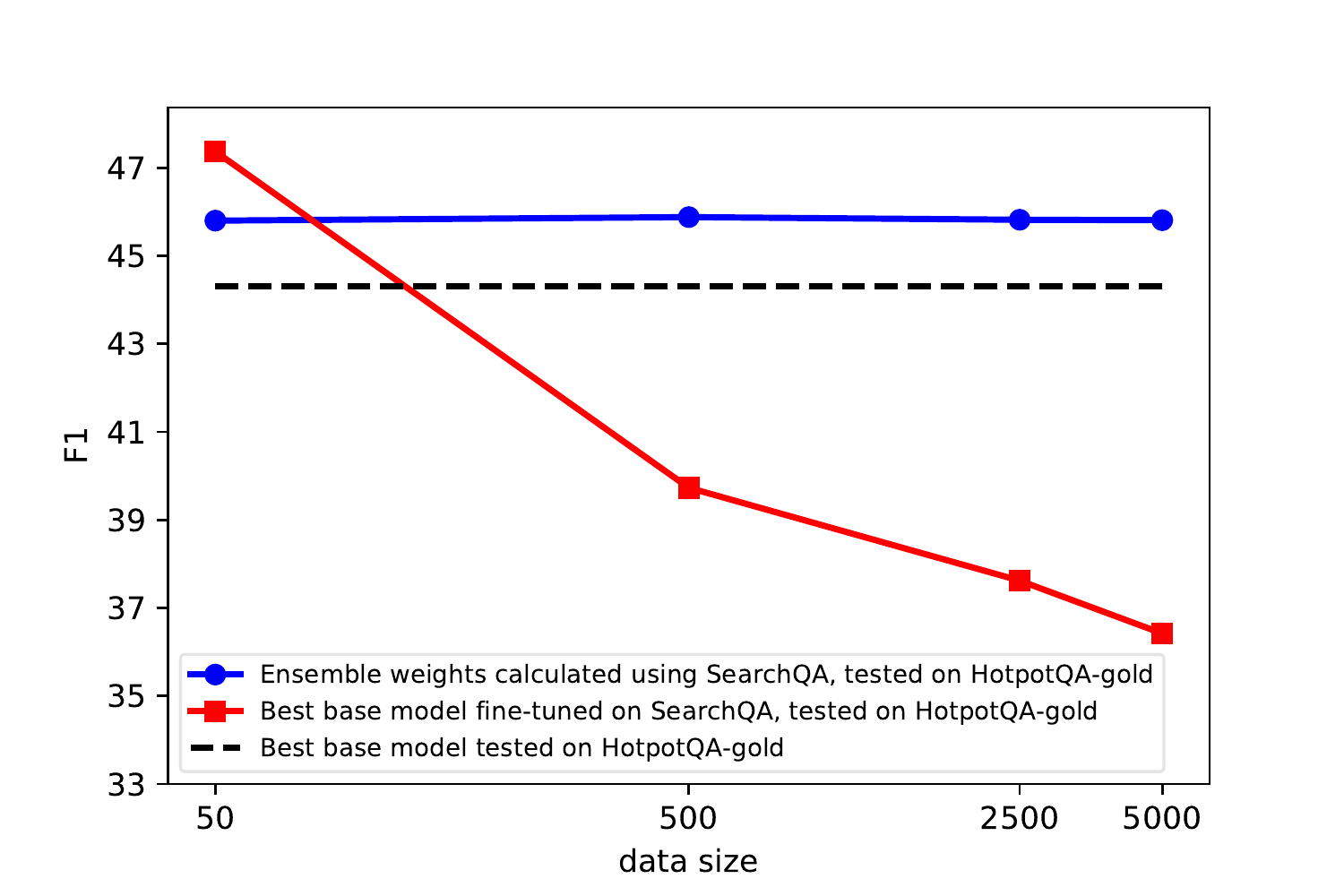} 
\end{subfigure}
\begin{subfigure}[b]{.45\linewidth}
\includegraphics[width=\linewidth,valign=T]{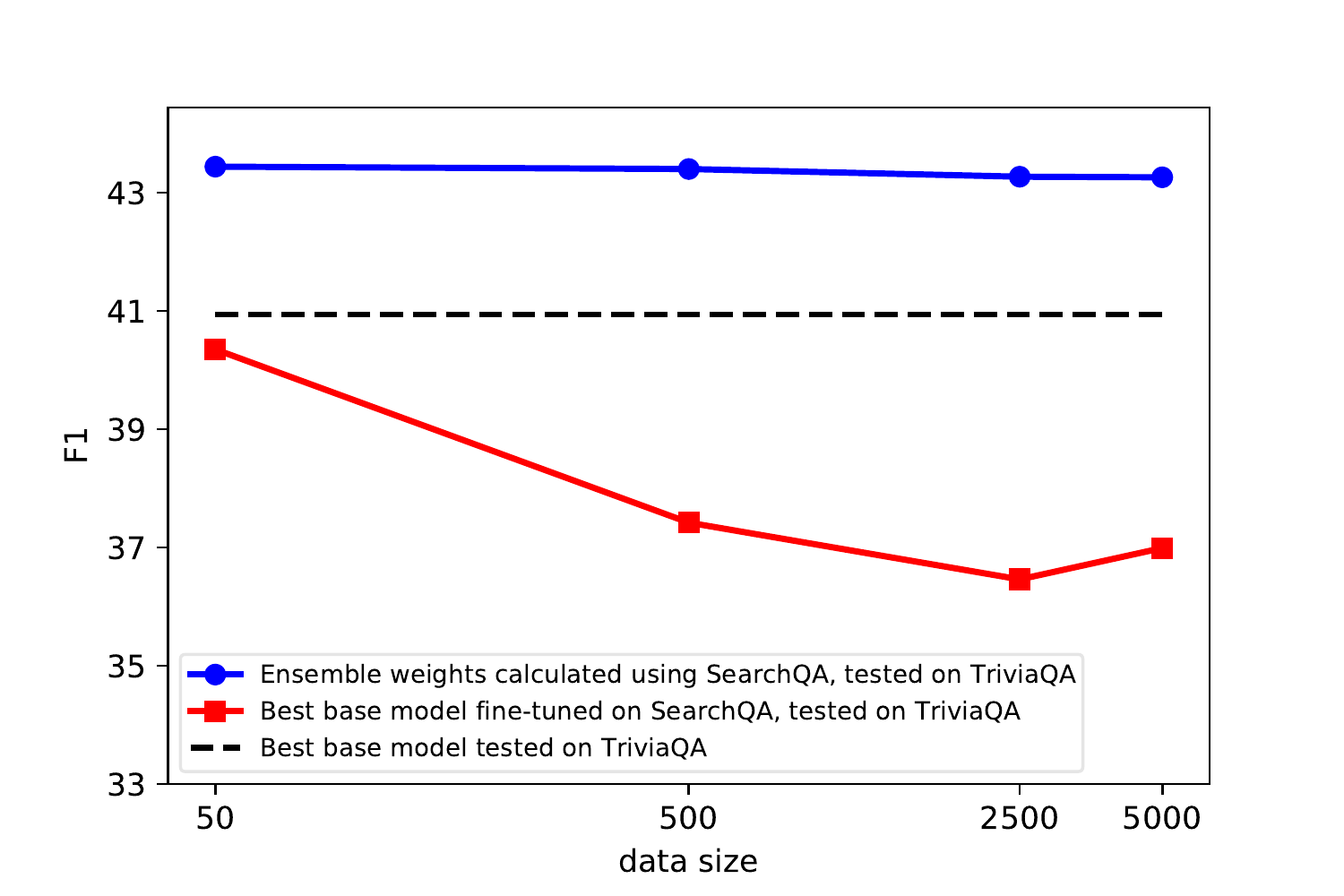} 
\end{subfigure}
\begin{subfigure}[b]{.45\linewidth}
\includegraphics[width=\linewidth,valign=T]{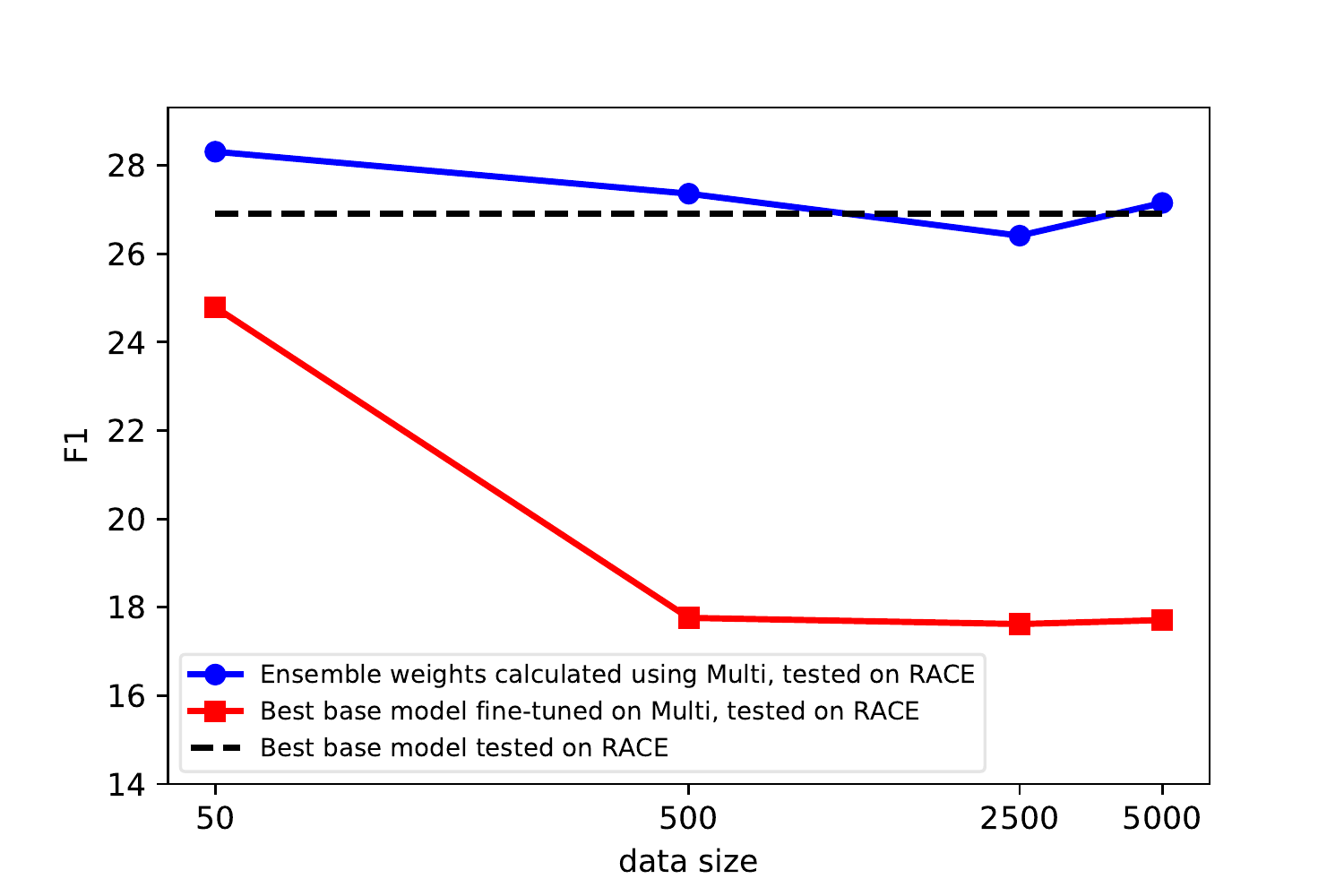} 
\end{subfigure}
\begin{subfigure}[b]{.45\linewidth}
\includegraphics[width=\linewidth,valign=T]{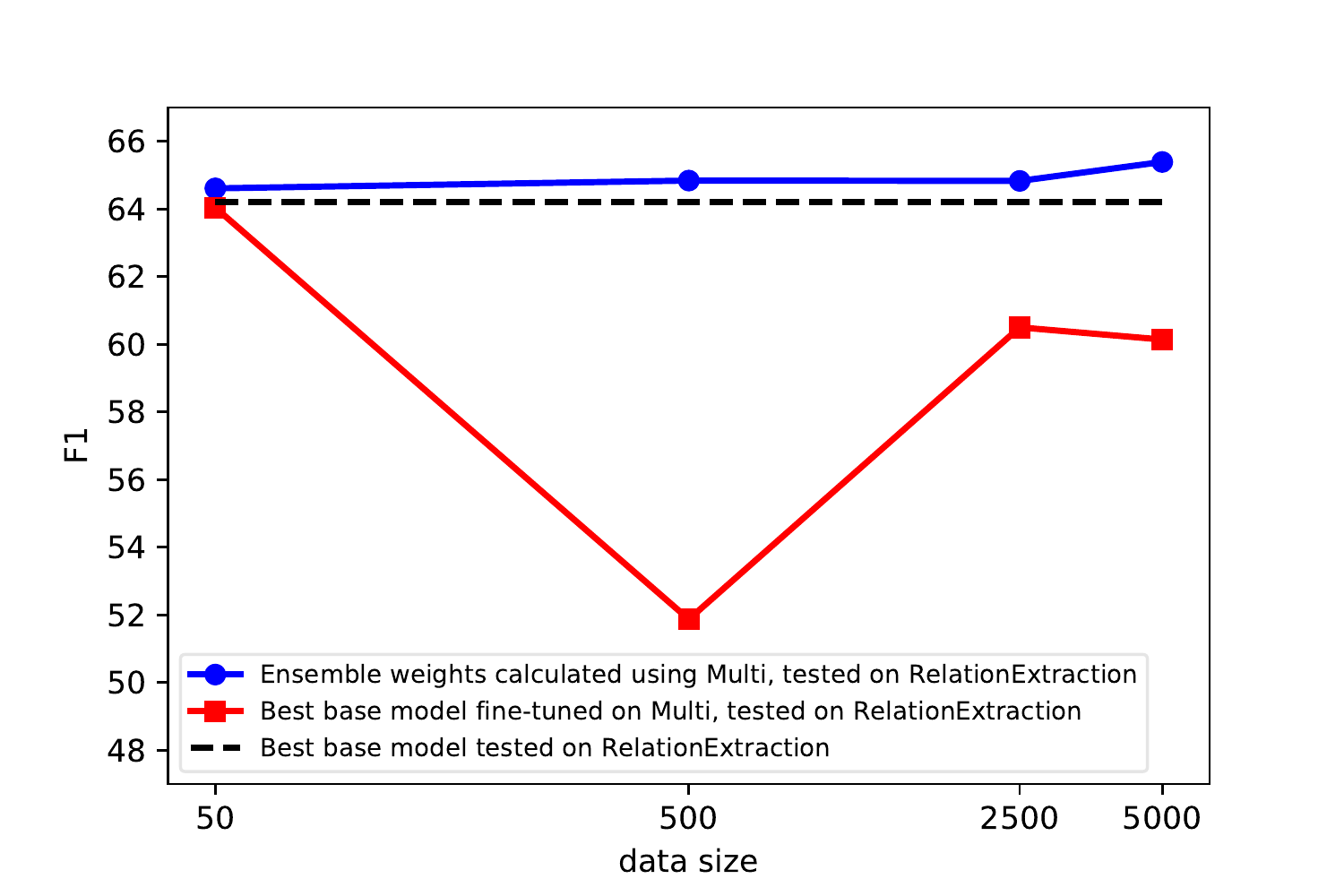} 
\end{subfigure}
\begin{subfigure}[b]{.45\linewidth}
\includegraphics[width=\linewidth,valign=T]{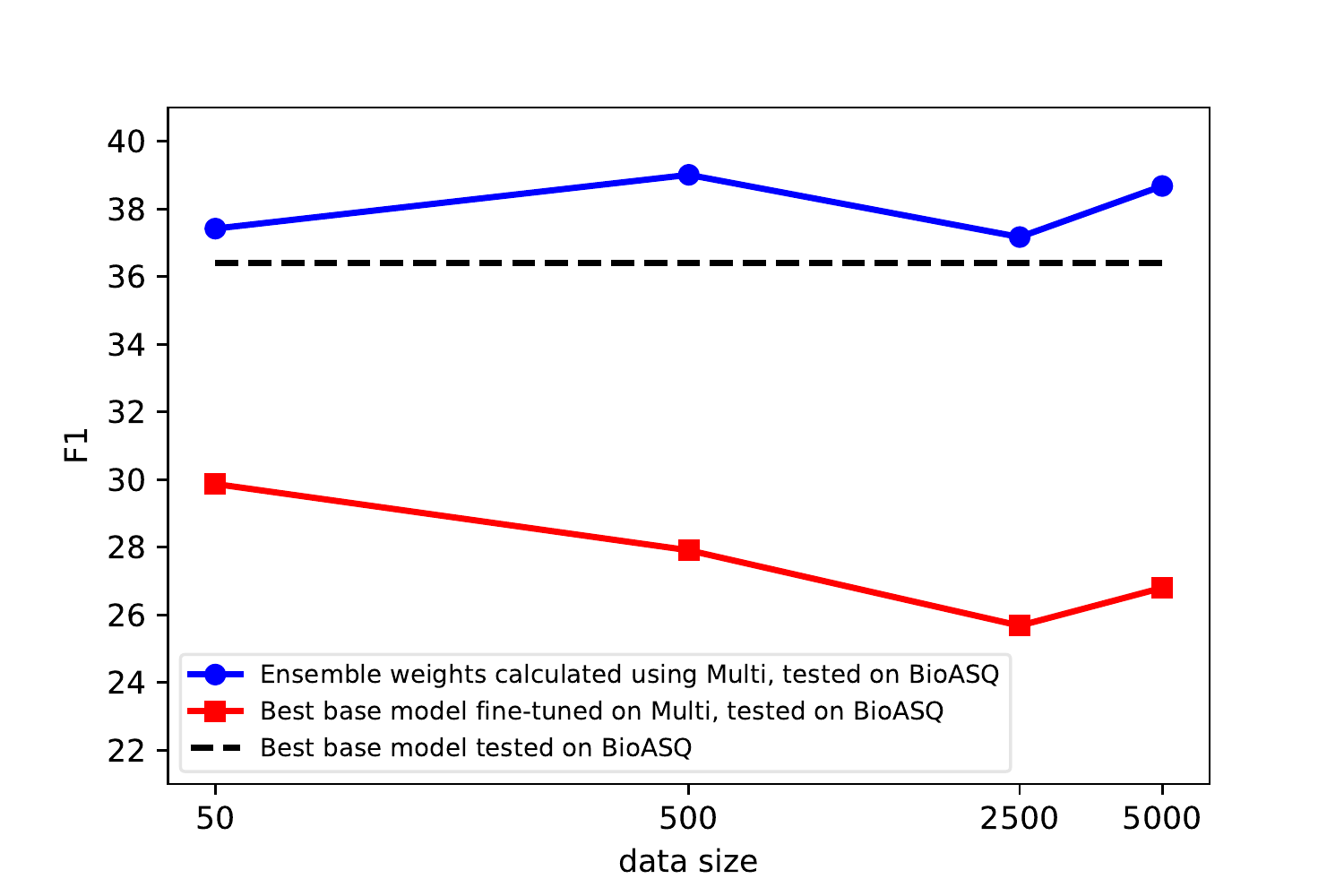} 
\end{subfigure}
\begin{subfigure}[b]{.45\linewidth}
\includegraphics[width=\linewidth,valign=T]{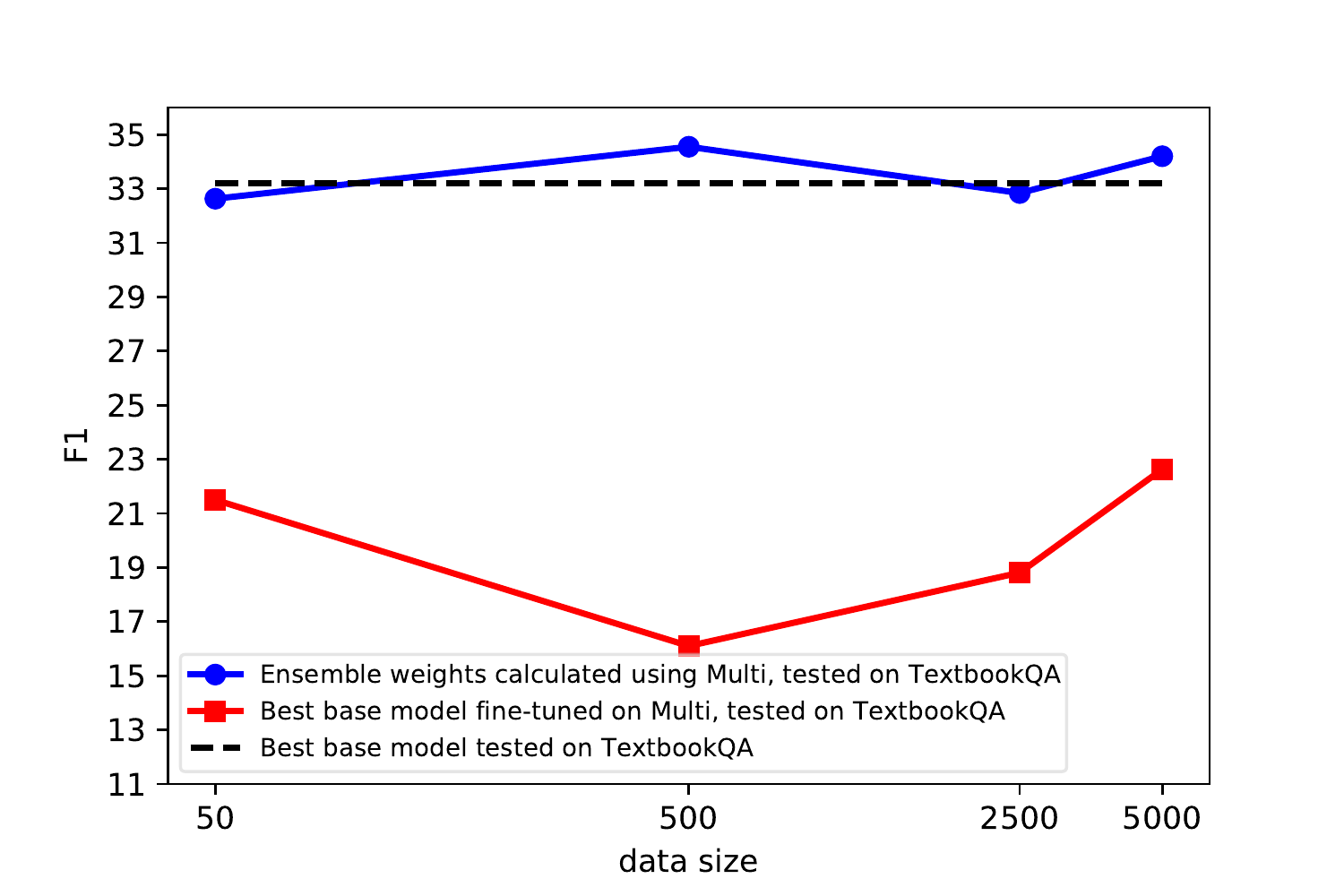} 
\end{subfigure}
\caption{Out-of-distribution generalization capability of the ensemble and fine-tuning approaches at different weight-estimation/fine-tuning set sizes. \textit{Multi} refers to an equal combination of HotpotQA-gold, SearchQA, and TriviaQA datasets.}
\label{fig:generalization}
\end{figure}

Finally, we explore the effect of different sizes of weight-estimation/fine-tuning set on the accuracy of ensemble and fine-tuning approaches for out-of-distribution data. This set is used for determining the weights of base models in the unequal weighting ensemble approach as well as for fine-tuning a base model. The used base models are as in the two previous experiments (Tables~\ref{tab-group2}~and~\ref{tab-group3}). Figure~\ref{fig:generalization} shows the obtained F1 values. According to this figure, the ensemble approach is (almost always) more accurate and more stable than the fine-tuning approach in generalizing to out-of-distribution data. The reason for the success of ensemble approach compared to fine-tuning in out-of-distribution generalization seems to be its lack of direct dependence to the distribution of the newly used data, as opposed to the fine-tuning approach which modifies the learned model to (over)fit the new dataset.
\begin{table}[!htbp]
\begin{center}
\begin{tabular}{l c c c c}
\hline
\diagbox{\textbf{Model}}{\textbf{Test Set}}&RACE&TextbookQA&BioASQ&RelationExtraction\\
\hline
BiDAF-SQuAD&26.90&33.19&36.40&64.21\\
BiDAF-NewsQA&21.75&23.03&24.58&44.96\\
BiDAF-Natural Questions&15.32&21.36&16.93&39.77\\
BiDAF-DROP&6.47&3.24&10.51&13.97\\
BiDAF-DuoRC&9.93&14.39&8.40&23.18\\
\hline
fine-tune-multi&17.71&22.63&26.80&60.14\\
\hline
ensemble-multi&\textbf{27.15}&\textbf{34.69}&\textbf{38.68}&\textbf{65.39}\\
\hline
\end{tabular}
\end{center} 
\caption{F1 score of base models, fine-tuning approach, and unequal weighting ensemble method. The fine-tuning and weight calculation are performed using a combination of HotpotQA-gold, SearchQA, and TriviaQA datasets; while the test is done on four completely unseen datasets.}
\label{tab-group3}
\end{table} 

\section{Conclusion}
The common paradigm in the natural language processing community to develop models for a new benchmark is to either train a new model or fine-tune a pre-trained one. In addition to their high computational costs and environmental effects~\citep{RN212}, their accuracies drop significantly for a new data distribution~\citep{talmor-berant-2019-multiqa}. 
In this paper, we investigated the effect of light-weight ensemble-based approach on the generalization of machine reading comprehension models to out-of-distribution data. The experiments were conducted using eight different datasets, six MRC models, and three settings including heterogeneous (different base models with the same training dataset), homogeneous (the same base model trained on different datasets), and hybrid. 

Different ensemble approaches were examined including weighting and stacking with probabilistic and non-probabilistic inputs. For the weighting approach, the unsupervised equal weighting and supervised accuracy-based unequal weighting were explored, where we introduced a simple dynamic method for exponentiation of base models' accuracies to be used as the weights. For the stacking method, we proposed a new architecture which obtained better or comparable results to other ensemble techniques. We also analyzed the obtained results in different settings to identify the reasons behind the success of ensemble techniques. 

Our experiments showed that with only the base models' outputs and without the need to retrain a new big model, we can obtain better out-of-distribution accuracies than the base MRC models. Analyzes indicated that low variance of the base models' accuracies in the target distribution and high variance of their outputs would have a positive effect on the performance of ensemble method. As an example, training a single model on different datasets obtained more diverse outputs than training different models on the same dataset, leading to a better model pool for the ensemble. 
We finally showed that the ensemble approach obtained better and more robust out-of-distribution generalization than the fine-tuning approach because of its lack of direct dependence to the distribution of the newly used data. 

As a future work, we are developing new algorithms for taking the input question and context features into account to measure the compatibility of the given input to each base model's dataset in a shared representation space. 

\bibliography{EL_based}
\clearpage
\appendix
\section{}

\setcounter{table}{0} 


\begin{table}[!htbp]
\scriptsize
\begin{subtable}[t]{.5\linewidth}
\centering
\begin{tabular}[t]{ m{2.5cm} I I}
\hline  \multirow{2}{*}{\diagbox{\textbf{Model}}{\textbf{Test Set}}} & \multicolumn{2}{G}{\textbf {HotpotQA-gold}} \\\cline{2-3} 
  & \textbf{EM} & \textbf{F1} \\
\hline
DrQA-SQuAD&24.32&35.85\\
BiDAF-SQuAD&31.18&44.30\\
QANet-SQuAD&32.06&46.26\\
DistilBERT-SQuAD&35.86&54.45\\
BERT-SQuAD&51.08&68.82\\
\hline
Mean-Dr-QA-DB-BE&42.59&58.93\\
Mul-Dr-QA-DB-BE&43.74&60.19\\
Mean-Dr-Bi-QA-BE&40.31&54.70\\
Mul-Dr-Bi-QA-BE&41.76&56.79\\
Mean-Dr-Bi-DB-BE&43.50&59.30\\
Mul-Dr-Bi-DB-BE&44.23&60.22\\
Mean-Bi-QA-DB-BE&43.45&59.49\\
Mul-Bi-QA-DB-BE&44.45&60.67\\
Mean-Dr-Bi-QA-DB&36.06&51.00\\
Mul-Dr-Bi-QA-DB&37.67&52.59\\\hline
Mean-Dr-Bi-QA&32.01&45.52\\
Mul- Dr-Bi-QA&32.96&46.66\\
Mean-Bi-QA-DB&35.60&50.58\\
Mul-Bi-QA-DB&37.45&52.70\\
Mean-Bi-QA-BE&41.09&55.65\\
Mul-Bi-QA-BE&43.38&58.50\\
Mean-QA-DB-BE&43.55&60.55\\
Mul-QA-DB-BE&44.84&61.93\\
Mean-Bi-DB-BE&44.86&61.17\\
Mul-Bi-DB-BE&45.30&62.07\\\hline
Mean-Bi-QA&33.50&47.36\\
Mul-Bi-QA&34.37&48.71\\
Mean-DB-BE&44.59&63.41\\
Mul-DB-BE&44.82&64.10\\
Mean-QA-DB&36.20&52.32\\
Mul-QA-DB&37.74&54.27\\
Mean-QA-BE&45.81&61.26\\
Mul-QA-BE&46.82&63.13\\
Mean-Bi-DB&36.79&52.39\\
Mul-Bi-DB&38.18&53.88\\\hline
Mean-All&40.89&56.20\\
Mul-All&41.89&57.53\\\hline
\hline
\end{tabular}
\caption{Heterogeneous}
\label{tab:A-g1-heter}
\end{subtable}
\begin{subtable}[t]{.5\linewidth}
\begin{tabular}[t]{ m{2.9cm} I I}
\hline  \multirow{2}{*}{\diagbox{\textbf{Model}}{\textbf{Test Set}}} & \multicolumn{2}{G}{\textbf {HotpotQA-gold}} \\\cline{2-3} 
  & \textbf{EM} & \textbf{F1} \\
\hline
 BiDAF-SQuAD & 31.18 &44.30 \\
 BiDAF-NewsQA & 22.55 & 37.31 \\
 BiDAF-Natural Questions & 11.34 & 20.10  \\
 BiDAF-DROP & 05.49 & 12.28 \\
 BIDAF-DuoRC & 20.20 & 29.63 \\
 \hline
 Mean-Ne-NQ-DR-Du&22.49&34.07\\
Mul-Ne-NQ-DR-Du&24.47&37.54\\
Mean-SQ-Ne-NQ-Du&30.66&44.13\\
Mul-SQ-Ne-NQ-Du&29.82&43.99\\
Mean-SQ-NQ-DR-Du&26.99&38.47\\
Mul-SQ-NQ-DR-Du&26.40&39.57\\
Mean-SQ-Ne-NQ-DR&27.71&40.78\\
Mul-SQ-Ne-NQ-DR&27.77&41.63\\
Mean-SQ-Ne-DR-Du&29.88&43.26\\
Mul-SQ-Ne-DR-Du&30.01&43.72\\\hline
Mean-Ne-DR-Du&20.74&32.58\\
Mul-Ne-DR-Du&19.64&31.66\\
Mean-NQ-DR-Du&14.20&22.90\\
Mul-NQ-DR-Du&15.47&25.99\\
Mean-SQ-Ne-NQ&25.77&38.35\\
Mul-SQ-Ne-NQ&25.10&38.57\\
Mean-SQ-Ne-Du&30.25&43.79\\
Mul-SQ-Ne-Du&28.32&42.38\\
Mean-Ne-NQ-Du&21.61&33.35\\
Mul-Ne-NQ-Du&21.96&34.82\\\hline
Mean-SQ-Ne & 30.62 & 45.18 \\
Mul-SQ-Ne & 30.86 & 45.86\\
Mean-Ne-Du & 25.18 & 39.21 \\
Mul-Ne-Du & 25.62 & 40.\\
Mean-SQ-Du & 31.28 & 44.01 \\
Mul-SQ-Du & 32.01 & 45.28 \\
Mean-NQ-DR& 11.00 & 19.24\\
Mul-NQ-DR&12.76&23.19\\
Mean-NQ-Du&19.96&32.35\\
Mul-NQ-Du&18.74&28.74\\\hline
Mean-All&29.59&42.41\\
Mul-All&26.27&39.04\\\hline
\hline
\end{tabular}
\caption{Homogeneous}
\label{tab:A-g1-homo}
\end{subtable}
\pagebreak

\end{table}
\begin{table}[ht!]\ContinuedFloat
\scriptsize
\begin{subtable}[t]{\textwidth}
\centering
\begin{tabular}[t]{ m{3.5cm} I I}
\hline  \multirow{2}{*}{\diagbox{\textbf{Model}}{\textbf{Test Set}}} & \multicolumn{2}{G}{\textbf {HotpotQA-gold}} \\\cline{2-3} 
  & \textbf{EM} & \textbf{F1} \\
\hline
 BiDAF-NewsQA&22.55&37.31\\
DrQA-SQuAD&24.32&35.85\\
NaQANet-DROP&12.25&20.79\\
QANet-Natural Questions&18.95&31.01\\
distilBERT-TriviaQA&20.98&32.23\\\hline
Mean-Bi-Dr-QA-DB&28.96&42.88\\
Mul-Bi-Dr-QA-DB&32.86&47.10\\
Mean-Bi-Dr-Na-QA&24.81&37.33\\
Mul- Bi-Dr-Na-QA&22.83&34.90\\
Mean-Bi-Dr-Na-DB&27.72&41.40\\
Mul-Bi-Dr-Na-DB&30.16&44.42\\
Mean-Bi-Na-QA-DB&25.77&39.51\\
Mul-Bi-Na-QA-DB&29.23&43.67\\
Mean-Dr-Na-QA-DB&26.28&38.73\\
Mul-Dr-Na-QA-DB&29.96&44.08\\\hline
Mean-Bi-Dr-DB&26.54&40.22\\
Mul-Bi-Dr-DB&29.08&43.16\\
Mean-Bi-Dr-QA&26.79&40.39\\
Mul-Bi-Dr-QA&27.08&40.09\\
Mean-Dr-QA-DB&26.03&38.55\\
Mul-Dr-QA-DB&29.93&43.70\\
Mean-Bi-QA-DB&24.81&39.08\\
Mul-Bi-QA-DB&27.30&41.86\\
Mean-Bi-Dr-Na&25.10&38.04\\
Mul-Bi-Dr-Na&24.28&36.86\\\hline
Mean-Bi-Dr&27.25&41.55\\
Mul-Bi-Dr&28.66&42.78\\
Mean-Bi-DB&24.72&39.34\\
Mul-Bi-DB&26.57&41.94\\
Mean-Dr-BE&25.84&39.05\\
Mul-Dr-BE&28.99&44.01\\
Mean-Dr-QA&25.93&38.65\\
Mul-Dr-QA&27.71&40.93\\
Mean-Dr-Na&23.69&34.97\\
Mul-Dr-Na&24.18&36.18\\\hline
Mean-All&27.81&41.06\\
Mul-All&30.74&44.55\\\hline
\hline
\end{tabular}
\caption{Hybrid}
\label{tab:A-g1-hybrid}
\end{subtable}
\caption{The extended version of Table~\ref{tab:g1} for investigating different combinations of base models in different evaluation settings. The used ensemble methods are the simple mean and multiplication (Eq.~\ref{eq:simple}) that are referred to as ``Mean'' and ``Mul'', respectively.}
\label{tab:A-g1}
\end{table}
\clearpage
\setcounter{figure}{0} 
\begin{figure}[!ht]
\centering
\begin{subfigure}[b]{.47\linewidth}
\includegraphics[width=\textwidth,valign=T]{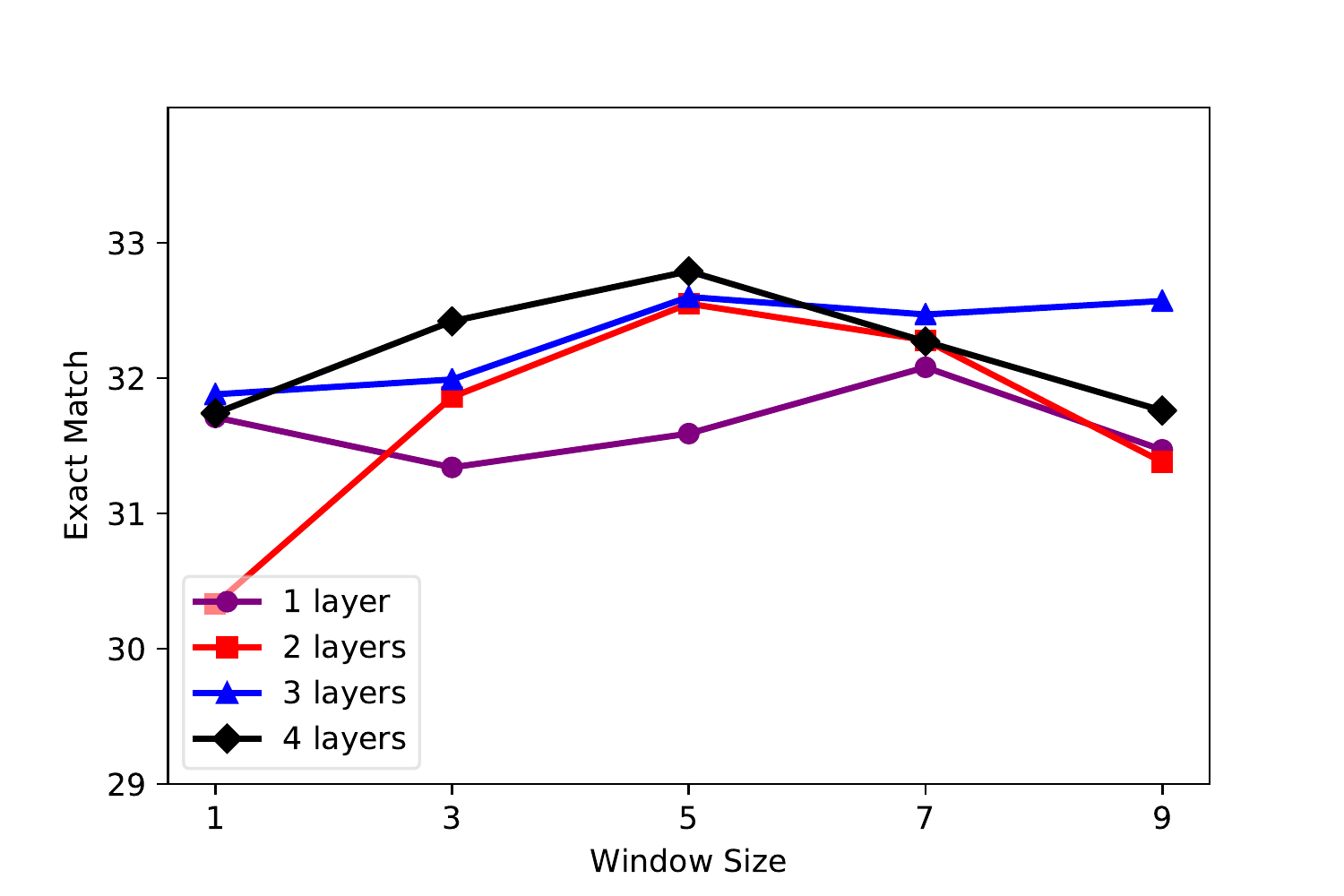} 
\end{subfigure}
\begin{subfigure}[b]{.47\linewidth}
\includegraphics[width=\linewidth,valign=T]{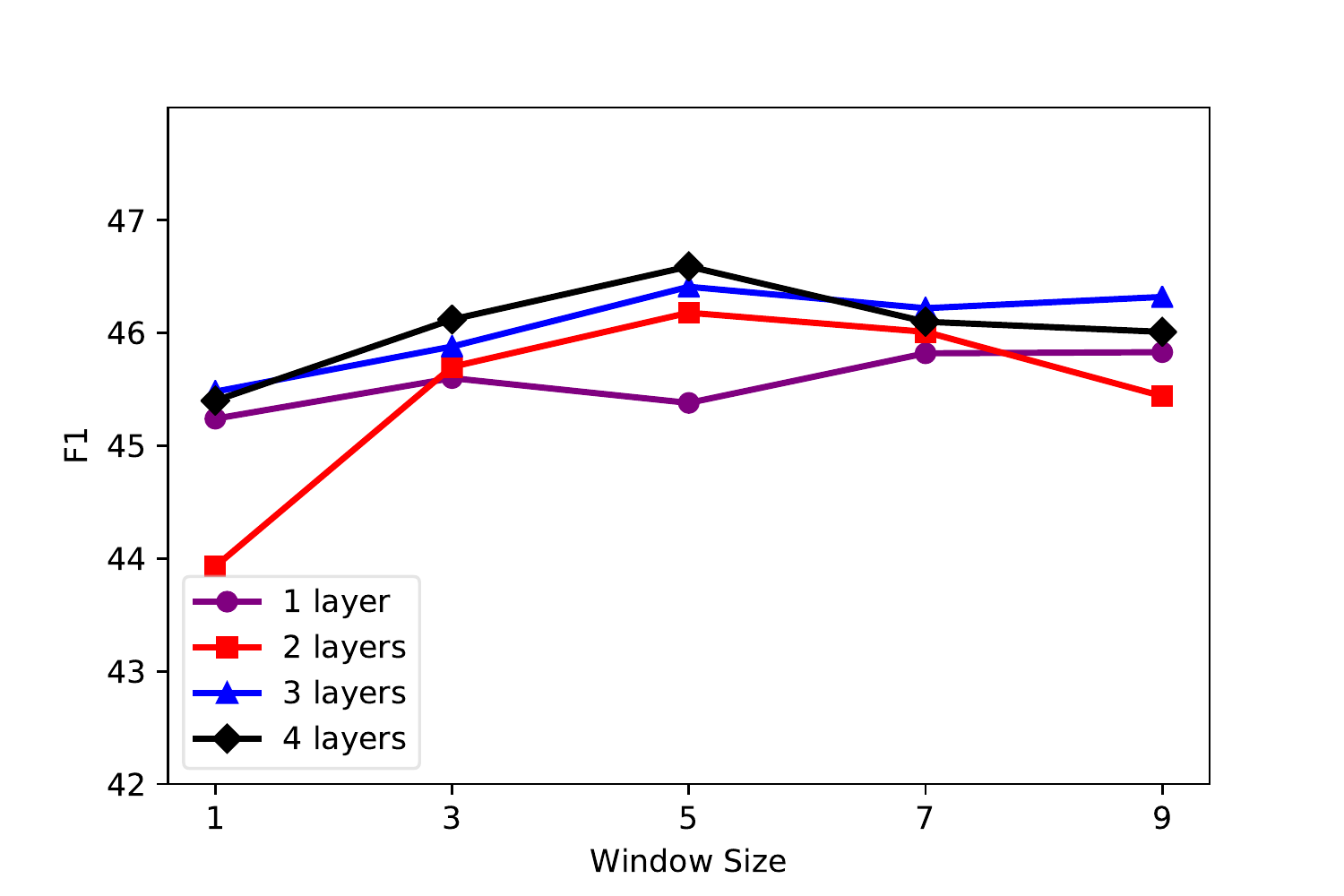} 
\end{subfigure}
\caption{Evaluating the number of layers and the window size in the stacking-based ensemble architecture. The BiDAF model is trained on SQuAD, NewsQA, Natural Questions, DROP, and DuoRC datasets to obtain the base models. The HotpotQA-gold dataset is used as the test set.}
\label{fig:A-stackingStructures}
\end{figure}

\end{document}